\documentclass{CVM}
\usepackage{comment}
\usepackage[utf8]{inputenc}

\definecolor{best}{rgb}{0.96, 0.57, 0.58}
\definecolor{second}{rgb}{0.98, 0.78, 0.57}
\definecolor{third}{rgb}{1.0, 1.0, 0.56}
\usepackage[table]{xcolor}
\usepackage{adjustbox}

\CVMsetup{
type      = {Research/Review Article},
doi       = {s41095-0xx-xxxx-x},
title     = {Monocular Depth Estimation from a Single Image: Progress and Opportunities},
author    = {Muxin Liu$^{1,\ast}$, Xiaoyang Lyu$^{1,\ast}$, Yang-Tian Sun$^{1,\ast}$, Yi-Hua Huang$^{1}$, Ziyi Yang$^{1}$, Peng Dai$^{1}$ and Xiaojuan Qi$^{1}$\cor{}\\
},
runauthor = {M. Liu, X. Lyu, Y.-T. Sun, Y.-H. Huang, Z. Yang, P. Dai, X. Qi},
abstract  = {
Monocular depth estimation has long stood as a fundamental challenge in computer vision, enabling a wide range of applications including 3D reconstruction, robotics, autonomous driving, and augmented reality. This survey traces the field’s evolution from early learning-based methods to the emergence of transformative foundation models. We begin by framing the problem, distinguishing between relative and metric depth estimation, and highlighting the key challenges that have shaped a decade of research.
We then present common problem formulations and introduce the most widely used datasets, covering indoor, outdoor, and synthetic data. Following this, we review major advances prior to the foundation model era, distilling core insights from influential methods that contributed to improvements in accuracy, efficiency, and robustness.
The survey then turns to the recent surge of foundation-model-based approaches, categorizing them into discriminative and generative paradigms and emphasizing the critical roles of large-scale pretraining (e.g., DINOv3) and synthetic data. We compare representative models using both quantitative benchmarks and qualitative examples, and discuss natural extensions to video-based depth estimation.
Further, to illustrate real-world impact, we highlight the integration of depth estimation into applications such as visual SLAM, content generation, and robot perception. Finally, we outline open challenges and promising research directions as the field advances further into the era of foundation models. Github link: \href{https://github.com/CVMI-Lab/Depth_Survey}{CVMI-Lab/Depth-Survey}.

},
keywords  = {Monocular Depth Estimation, Metric Depth, Relative Depth, Foundation Models, Vision Transformers, Diffusion Models},
copyright = {The Author(s)},
}

\address{Department of Electrical and Computer Engineering, The University of Hong Kong, Hong Kong, 999077. E-mail: Muxin Liu, mxliu@connect.hku.hk; Xiaoyang Lyu, shawlyu@connect.hku.hk; Yang-Tian Sun, sunyangtian98@gmail.com; Yi-Hua Huang, huangyihua16@mails.ucas.ac.cn; Ziyi Yang, 14ziyiyang@gmail.com; Peng Dai, daipeng@eee.hku.hk; Xiaojuan Qi, xjqi@eee.hku.hk\cor{}.}

\begin{document}

\maketitle



\section{Introduction and Background}
Depth information is crucial for spatial perception in applications such as robotics, autonomous driving, 3D modeling, virtual and augmented reality, and computational photography. Traditional methods rely on multiple images or specialized sensors to capture depth, such as stereo cameras and Kinect~\cite{scharstein2002taxonomy}. However, these approaches increase hardware complexity and costs. In contrast, monocular depth estimation offers greater flexibility and has garnered significant attention.

Early learning-based attempts to predict depth from a single image date back over a decade. Prior to deep learning, Saxena \textit{et al.}~\cite{Saxena2008Make3DDP} pioneered data-driven monocular depth prediction using Markov Random Fields on hand-crafted features. The breakthrough came with the advent of deep convolutional neural networks. Eigen \textit{et al.}~\cite{eigen2014depth} demonstrated that a multi-scale CNN could directly regress a dense depth map from a single RGB input, kickstarting the modern era of monocular depth estimation (MDE). Subsequent works rapidly improved accuracy on benchmark datasets by introducing deeper networks~\cite{kundu2018adadepth,mayer2016large,laina2016deeperdepthpredictionfully}, better loss functions (e.g. scale-invariant or ordinal regression losses), and leveraging multi-task cues such as surface normals~\cite{qi2018geonet,qi2020geonet++}, optical flow~\cite{yin2018geonet,saxena2023surprising}, or semantics~\cite{eigen2015predicting}.

Initially, monocular depth estimation models struggled with poor generalization when tested outside their training domains. For instance, a model trained on indoor scenes from the NYU Depth v2 dataset~\cite{Silberman2012IndoorSA} would fail when tested on outdoor driving scenes like KITTI~\cite{geiger2013vision}, and vice versa, due to differences in scene scale and camera intrinsics. Additionally, models that achieved state-of-the-art accuracy on specific datasets often learned dataset-specific biases (e.g., limited depth range or field of view), preventing them from generalizing to arbitrary images.

In recent years, the field has seen the rise of \emph{foundation models} for depth estimation~\cite{xu2026towards}, following the trend established in natural language processing. Depth foundation models are large models trained on vast and diverse datasets (including unlabeled or synthetic data) to achieve strong zero-shot performance on a wide range of in-the-wild images. The aim is to create a single model that is broadly applicable, rather than training specialized models for each dataset. Early efforts in this direction for depth estimation included combining multiple training datasets with carefully designed loss function design~\cite{ranftl2020towards} and employing vision transformers to better capture global context and enhance learning capabilities~\cite{ranftl2021vision}. The latest foundation models have advanced even further, leveraging tens of millions of images~\cite{yang2024depth} to learn generalizable knowledge for depth predictions. These models are often orders of magnitude more data-hungry and parameter-rich than traditional methods, resulting in dramatic improvements in zero-shot generalization.

\begin{figure*}[htbp]
    \centering
    \includegraphics[width=1.02\linewidth]{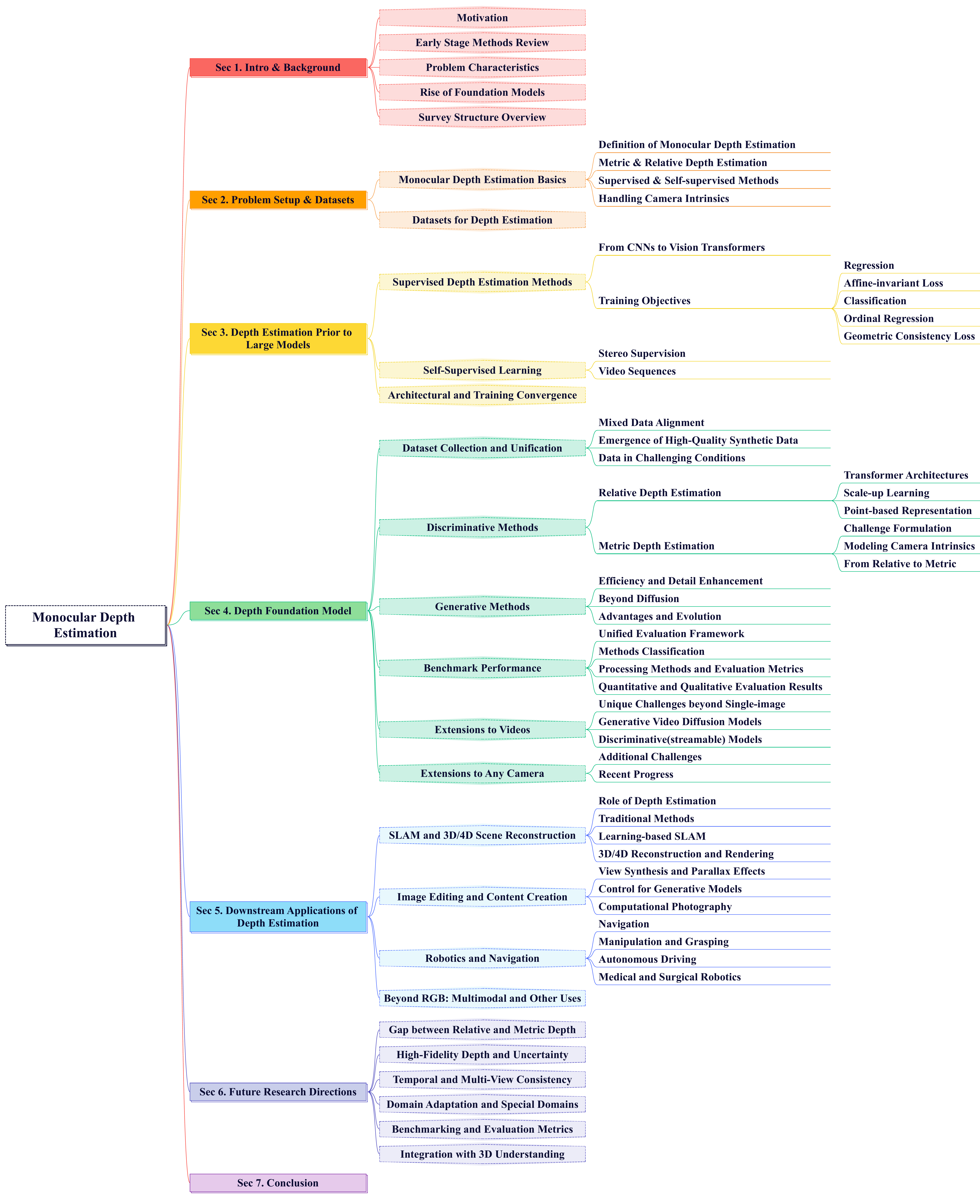}
    \caption{The survey is structured as follows: we begin with the background and datasets of monocular depth estimation (MDE) ), followed by a review of methods developed prior to the advent of foundation models. We then examine foundational depth estimation approaches and conclude with an overview of key downstream applications of MDE and future research directions.}
    \label{fig:mindmap}
\end{figure*}

This survey is organized as follows (see Fig.~\ref{fig:mindmap}). 
\textbf{Section~\ref{sec:background}} formalizes the problem setup and provides an overview of commonly used datasets for monocular depth estimation. 
\textbf{Section~\ref{sec:depth prior}} reviews key developments prior to the foundation model era, encompassing both supervised and self-supervised learning approaches, and highlights architectural innovations and the design of effective loss functions. 
\textbf{Section~\ref{sec:Depth Foundation Model}} focuses on the emergence of depth foundation models, which constitutes the core contribution of this survey. 
We examine representative methods from three perspectives: (a) dataset collection and unification; (b) discriminative models; and (c) generative models. 
{To address the inconsistency of evaluation protocols across existing works, we introduce a unified evaluation pipeline and conduct systematic quantitative and qualitative comparisons on representative models, enabling fair benchmarking and clearer insights into recent advances.} 
We further discuss extensions of these models to video-based depth estimation. 
\textbf{Section~\ref{sec:downstream}} surveys the application of monocular depth estimation in downstream tasks, including SLAM and 3D reconstruction, image editing and content creation, and robotics. 
Finally, \textbf{Section~\ref{sec:future}} outlines open challenges and future directions, such as improving fine-grained detail and temporal consistency, supporting diverse camera models, and incorporating geometric priors.

\section{Problem Setup and Datasets}
\label{sec:background}
\subsection{Monocular Depth Estimation Basics}

Monocular depth estimation aims to predict a dense depth map $D$ from a single RGB image $I$, where each depth value 
$D(u,v)$ represents the distance from the camera to the scene along the viewing ray of pixel  $(u,v)$.
  
\begin{figure*}[htbp]
    \centering
    \includegraphics[width=0.95\linewidth]{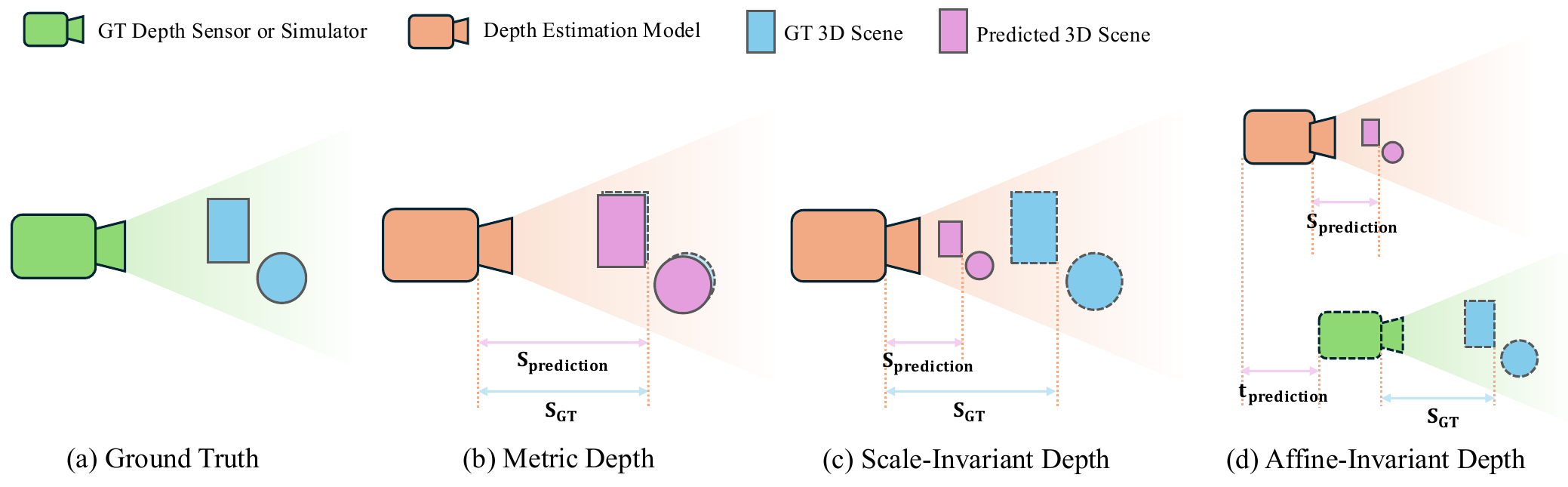}
    \caption{Illustration of different depth estimation paradigms: (a) Ground-truth depth acquired from a depth sensor or simulator; (b) Metric depth estimation, the predicted depth is consistent with the ground truth in physical units; (c) Scale-invariant depth estimation, the predicted depth differs from the ground truth by a global scaling factor; (d) Affine-invariant depth estimation, the predicted depth diverges from the ground truth by both a global scale and an additive translation.}
    \label{fig:metricandrelative}
\end{figure*}

As illustrated in Fig.~\ref{fig:metricandrelative}, there are two main types of depth estimation:
\begin{itemize} 
\item \textbf{Metric depth estimation}~\cite{wang2025moge2accuratemonoculargeometry, piccinelli2024unidepth, piccinelli2025unidepthv2universalmonocularmetric, bochkovskii2024depth, bhat2023zoedepth, guizilini2023zeroshotscaleawaremonoculardepth, yin2023metric3d, Hu_2024,liu2026foundationgeo} produces depth values in real-world units (e.g., meters), aligned with the true scale of the scene. 
\item \textbf{Relative depth estimation}~\cite{ke2024repurposing,wang2025moge,yang2024depth,depth_anything_v2,fu2024geowizard,yin2021learning,he2024lotus,birkl2023midas,ranftl2021vision,garcia2025fine,xu2024matters,liu2026foundationgeo}-- also known as scale-invariant depth estimation --predicts depth up to an unknown scale (and sometimes an additive shift). In this case, if $D$ is a valid prediction, then so is $\alpha D + \beta$ for any scale $\alpha>0$ (and a shift $\beta$ for affine-invariant depth estimation) is equally valid. Relative depth preserves the correct ordering and proportional relationships between depths but not their absolute magnitudes.
\end{itemize}

\begin{figure*}[!htb]
\begin{minipage}{0.5\linewidth}
\begin{tikzpicture}[scale=1]

\tikzstyle{every node}=[font=\tiny,scale=1]
    \fill[gray!60] (0.5,4) circle (1.2mm) node[black, anchor=south, yshift=1.5mm] {\small{2014}};
    \fill[gray!60] (2.2,4) circle (1.2mm) node[black, anchor=north, yshift=-1.5mm] {\small{2016}};
    \fill[gray!60] (3.9,4) circle (1.2mm) node[black, anchor=south, yshift=1.5mm] {\small{2017}};
    \fill[gray!60] (5.6,4) circle (1.2mm) node[black, anchor=north, yshift=-1.5mm] {\small{2018}};
    \fill[gray!60] (7.3,4) circle (1.2mm) node[black, anchor=south, yshift=1.5mm] {\small{2019}};
    \fill[gray!60] (9.0,4) circle (1.2mm) node[black, anchor=north, yshift=-1.5mm] {\small{2020}};
    \fill[gray!60] (10.7,4) circle (1.2mm) node[black, anchor=south, yshift=1.5mm]  {\small{2021}};
    \fill[gray!60] (12.4,4) circle (1.2mm) node[black, anchor=north, yshift=-1.5mm] {\small{2023}};
    \fill[gray!60] (14.1,4) circle (1.2mm) node[black, anchor=south, yshift=1.5mm]  {\small{2024}};
    \fill[gray!60] (15.6,4) circle (1.2mm) node[black, anchor=north, yshift=-1.5mm]  {\small{2025}};

    \draw[magenta, line width=0.5mm] (0.2, 7) -- (1.2, 7) node[black, anchor=west] {\small{Metric Depth}};
    \draw[cyan, line width=0.5mm] (4.2, 7) -- (5.2, 7) node[black, anchor=west] {\small{Relative Depth}};

    \draw[gray!60, line width=0.5mm] (0.5, 4) -- (0.5, 3) ;
    \draw[magenta, line width=0.5mm] (0.5, 3) -- (0.8, 2.8)
    node[anchor=north, text width=0cm, xshift=2mm, yshift=3mm] {\small{ \text{Eigen}~\etal~\cite{eigen2014depth} }};

    \draw[gray!60, line width=0.5mm] (2.2, 4) -- (2.2, 4.7);
    \draw[magenta, line width=0.5mm] (2.2, 4.7) -- (2.4, 4.9)
    node[anchor=south, text width=0cm, xshift=2mm, yshift=-3mm] {\small{\text{Monodepth}~\cite{godard2017unsupervisedmonoculardepthestimation} }};
    \draw[gray!60, line width=0.5mm] (2.2, 4) -- (2.2, 5.2);
    \draw[magenta, line width=0.5mm] (2.2, 5.2) -- (2.4, 5.4)
    node[anchor=south, text width=0cm, xshift=2mm, yshift=-3mm] {\small{\text{Garg}~\etal~\cite{garg2016unsupervisedcnnsingleview} }};
    \draw[gray!60, line width=0.5mm] (2.2, 4) -- (2.2, 5.7);
    \draw[magenta, line width=0.5mm] (2.2, 5.7) -- (2.4, 5.9)
    node[anchor=south, text width=0cm, xshift=2mm, yshift=-3mm] {\small{\text{Laina}~\etal~\cite{laina2016deeperdepthpredictionfully} }};

    \draw[gray!60, line width=0.5mm] (3.9, 4) -- (3.9, 3) ;
    \draw[magenta, line width=0.5mm] (3.9, 3) -- (4.1, 2.8)
    node[anchor=north, text width=0cm, xshift=2mm, yshift=3mm] {\small{ \text{SfM Learner}~\cite{zhou2017unsupervised} }};

    \draw[gray!60, line width=0.5mm] (5.6, 4) -- (5.6, 4.7);
    \draw[magenta, line width=0.5mm] (5.6, 4.7) -- (5.8, 4.9)
    node[anchor=south, text width=0cm, xshift=2mm, yshift=-3mm] {\small{\text{DORN}~\cite{fu2018deep} }};
    \draw[gray!60, line width=0.5mm] (5.6, 4) -- (5.6, 5.2);
    \draw[magenta, line width=0.5mm] (5.6, 5.2) -- (5.8, 5.4)
    node[anchor=south, text width=0cm, xshift=2mm, yshift=-3mm] {\small{\text{Monodepth2}~\cite{monodepth2} }};

    \draw[gray!60, line width=0.5mm] (7.3, 4) -- (7.3, 3) ;
    \draw[cyan, line width=0.5mm] (7.3, 3) -- (7.5, 2.8)
    node[anchor=north, text width=0cm, xshift=2mm, yshift=3mm] {\small{ \text{MiDaS}~\cite{ranftl2020towards} }};

    \draw[gray!60, line width=0.5mm] (9.0, 4) -- (9.0, 5.2);
    \draw[magenta, line width=0.5mm] (9.0, 5.2) -- (9.2, 5.4)
    node[anchor=south, text width=0cm, xshift=2mm, yshift=-3mm] {\small{\text{LeReS}~\cite{yin2021learning} }};

    \draw[gray!60, line width=0.5mm] (10.7, 4) -- (10.7, 2.5) ;
    \draw[cyan, line width=0.5mm] (10.7, 2.5) -- (10.9, 2.3)
    node[anchor=north, text width=0cm, xshift=2mm, yshift=3mm] {\small{ \text{DPT}~\cite{ranftl2021vision} }};
    \draw[gray!60, line width=0.5mm] (10.7, 4) -- (10.7, 3) ;
    \draw[cyan, line width=0.5mm] (10.7, 3) -- (10.9, 2.8)
    node[anchor=north, text width=0cm, xshift=2mm, yshift=3mm] {\small{ \text{Omnidata}~\cite{eftekhar2021omnidata} }};

    \draw[gray!60, line width=0.5mm] (12.4, 4) -- (12.4, 4.7);
    \draw[magenta, line width=0.5mm] (12.4, 4.7) -- (12.6, 4.9)
    node[anchor=south, text width=0cm, xshift=2mm, yshift=-3mm] {\small{\text{ZoeDepth}~\cite{bhat2023zoedepth} }};
    \draw[gray!60, line width=0.5mm] (12.4, 4) -- (12.4, 5.2);
    \draw[magenta, line width=0.5mm] (12.4, 5.2) -- (12.6, 5.4)
    node[anchor=south, text width=0cm, xshift=2mm, yshift=-3mm] {\small{\text{Metric3D}~\cite{yin2023metric3d} }};
    \draw[gray!60, line width=0.5mm] (12.4, 4) -- (12.4, 5.7);
    \draw[cyan, line width=0.5mm] (12.4, 5.7) -- (12.6, 5.9)
    node[anchor=south, text width=0cm, xshift=2mm, yshift=-3mm] {\small{\text{Marigold}~\cite{ke2024repurposing} }};

    \draw[gray!60, line width=0.5mm] (14.1, 4) -- (14.1, 2) ;
    \draw[magenta, line width=0.5mm] (14.1, 2) -- (14.3, 1.8)
    node[anchor=north, text width=0cm, xshift=2mm, yshift=3mm] {\small{ \text{UniDepth}~\cite{piccinelli2024unidepth} }};
    \draw[gray!60, line width=0.5mm] (14.1, 4) -- (14.1, 2.5) ;
    \draw[cyan, line width=0.5mm] (14.1, 2.5) -- (14.3, 2.3)
    node[anchor=north, text width=0cm, xshift=2mm, yshift=3mm] {\small{ \text{Depth Anything}~\cite{yang2024depth} }};
    \draw[gray!60, line width=0.5mm] (14.1, 4) -- (14.1, 3) ;
    \draw[cyan, line width=0.5mm] (14.1, 3) -- (14.3, 2.8)
    node[anchor=north, text width=0cm, xshift=2mm, yshift=3mm] {\small{ \text{DiffusionE2E}~\cite{garcia2025fine} }};
    \draw[gray!60, line width=0.5mm] (14.1, 4) -- (14.1, 3.5) ;
    \draw[cyan, line width=0.5mm] (14.1, 3.5) -- (14.3, 3.3)
    node[anchor=north, text width=0cm, xshift=2mm, yshift=3mm] {\small{ \text{Geowizard}~\cite{fu2024geowizard} }};

    \draw[gray!60, line width=0.5mm] (15.6, 4) -- (15.6, 4.5);
    \draw[magenta, line width=0.5mm] (15.6, 4.5) -- (15.8, 4.7)
    node[anchor=south, text width=0cm, xshift=2mm, yshift=-3mm] {\small{\text{Depth Pro}~\cite{bochkovskii2024depth} }};
    \draw[gray!60, line width=0.5mm] (15.6, 4) -- (15.6, 5.0);
    \draw[cyan, line width=0.5mm] (15.6, 5.0) -- (15.8, 5.2)
    node[anchor=south, text width=0cm, xshift=2mm, yshift=-3mm] {\small{\text{MoGe}~\cite{wang2025moge} }};
    \draw[gray!60, line width=0.5mm] (15.6, 4) -- (15.6, 5.5);
    \draw[magenta, line width=0.5mm] (15.6, 5.5) -- (15.8, 5.7)
    node[anchor=south, text width=0cm, xshift=2mm, yshift=-3mm] {\small{\text{MoGe-2}~\cite{wang2025moge2accuratemonoculargeometry} }};
    \draw[gray!60, line width=0.5mm] (15.6, 4) -- (15.6, 6.0);
    \draw[cyan, line width=0.5mm] (15.6, 6.0) -- (15.8, 6.2)
    node[anchor=south, text width=0cm, xshift=2mm, yshift=-3mm] {\small{\text{VGGT}~\cite{wang2025vggt} }};
    \draw[gray!60, line width=0.5mm] (15.6, 4) -- (15.6, 6.5);
    \draw[cyan, line width=0.5mm] (15.6, 6.5) -- (15.8, 6.7)
    node[anchor=south, text width=0cm, xshift=2mm, yshift=-3mm] {\small{\text{DAv3}~\cite{depthanything3} }};

    \draw[gray!60, -latex, line width=0.5mm] (0.2, 4) -- (17.5, 4);
\end{tikzpicture}

\end{minipage}
\caption{A brief timeline of representative works with their corresponding representations. Methods estimating metric depth are highlighted in red, while those estimating relative depth are marked in blue.}
\label{fig:timeline}
\end{figure*}
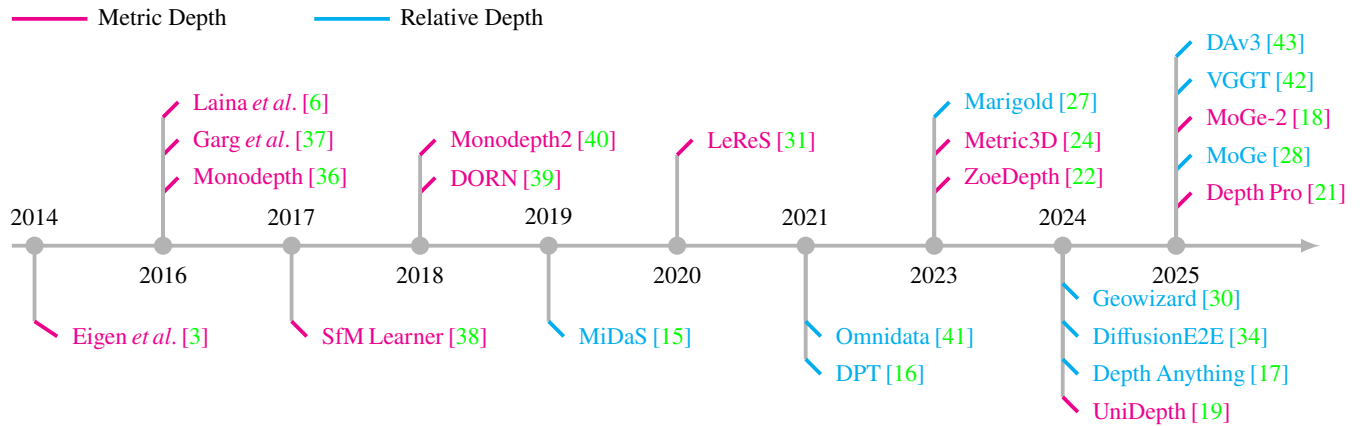

Monocular depth estimation is inherently ill-posed, since multiple 3D scenes with different scales can lead to identical 2D observations. Classical approaches~\cite{pentland1987new,nayar2002shape,hoiem2005automatic} addressed this ambiguity using handcrafted photometric and geometric cues, including shading, perspective, occlusion, and defocus, under restrictive assumptions. In contrast, modern learning-based methods resolve depth ambiguities by statistically learning such cues from large-scale data, resulting in improved robustness and generalization~\cite{Danier_2025_CVPR}. Supervised learning approaches rely on ground-truth depth maps, typically acquired using RGB-D sensors (e.g., structured light or time-of-flight cameras), LiDAR, or multi-view stereo for learning. In contrast, self-supervised methods~\cite{zhou2017unsupervised,monodepth2} estimate depth by enforcing geometric consistency across stereo pairs or video sequences, eliminating the need for explicit depth labels~\cite{mur2015orb}.

A particularly important challenge for metric depth estimation is the variation in camera intrinsics. Changes in focal length, for instance, directly affect the mapping between image content and absolute depth~\cite{yin2023metric3d}. Many early methods assume fixed camera intrinsics, for example, all training images share a known or narrow focal length range~\cite{eigen2014depth,eigen2015predicting,he2016deep,fu2018deep,He_2018,wang2019taskawaremonoculardepthestimation}. Recent advances have introduced techniques to generalize across unknown camera parameters, such as conditioning the model on camera intrinsics~\cite{piccinelli2024unidepth} or normalizing input images to a canonical camera model~\cite{yin2023metric3d}. These strategies will be discussed in more detail in the following sections.

\subsection{Datasets for Depth Estimation}

Progress in monocular depth estimation is closely tied to the availability of diverse datasets, which we categorize into four major types: synthetic, indoor, outdoor, and depth-in-the-wild datasets as summarized in Table~\ref{tab:datasets}. Each type offers unique benefits and poses specific limitations that influence model performance and generalization.
\begin{itemize}
\item \textbf{Synthetic datasets} such as TartanAir~\cite{tartanair2020iros}, FoundationStereo~\cite{wen2025stereo}, and BlendedMVS~\cite{yao2020blendedmvs} simulate highly photorealistic indoor and outdoor scenes with perfect ground-truth depth. Datasets like HyperSim~\cite{roberts:2021} and Replica~\cite{replica19arxiv} focus on indoor environments, while Virtual KITTI 2~\cite{cabon2020virtual} and MatrixCity~\cite{li2023matrixcity} target outdoor scenarios. They provide complete and perfectly labeled annotations under controlled conditions, but often exhibit a gap in realism and diversity when compared to real-world data. 
\item \textbf{Indoor datasets} such as NYU Depth V2~\cite{Silberman2012IndoorSA}, ARKitScenes~\cite{yin2023metric3d}, and SUN RGB-D~\cite{song2015sun} use RGB-D sensors to provide dense, short-range metric depth in residential or office environments. More recent benchmarks like, ScanNet++~\cite{yeshwanth2023scannethighfidelitydataset3d} improve annotation fidelity and scene diversity. These datasets offer short-range depth in indoor scenes.  
However, they often suffer from low resolution, and the inherent limitations of depth sensors can lead to missing regions or artifacts, particularly in reflective or transparent areas such as windows and mirrors.
\item \textbf{Outdoor datasets} like KITTI~\cite{geiger2013vision}, Nuscenes~\cite{nuscenes2019}, and Waymo~\cite{Sun_2020_CVPR} provide real-world driving scenes with long-range depth annotations obtained from LiDAR. More recent datasets such as Argoverse2~\cite{li2024patchfusion} and A2D2~\cite{saxena2024zero} further expand sensor coverage and scene complexity. However, LiDAR-based ground-truth is often sparse and noisy-- particularly for distant or dynamic objects-- and annotation quality varies across sensors and environmental conditions.
\item \textbf{Depth-in-the-wild datasets} like DIW~\cite{chen2017singleimagedepthperceptionwild}, MegaDepth~\cite{MegaDepthLi18}, and ReDWeb~\cite{Xian_2018_CVPR} use Internet imagery and SfM to provide diverse, large-scale annotations. They capture real-world diversity and help reduce dataset bias; however, the depth annotations are typically approximate-- such as ordinal or pseudo-depth-- and lack precise metric accuracy.
\end{itemize}

\begin{table*}[t]
\centering
\caption{\textbf{Overview of common datasets for monocular depth estimation.} Datasets are categorized into synthetic, indoor, outdoor/driving, and internet-based (``wild'') scenarios, detailing their size, resolution, ground-truth depth acquisition methods, and distinctive characteristics from official websites or introduction.}
\label{tab:datasets}
\renewcommand{\arraystretch}{1.1}
\resizebox{\linewidth}{!}{
\begin{tabular}{lcccccl}
\toprule
\textbf{Dataset} & \textbf{Type} & \textbf{Size} & \textbf{Resolution} &  \textbf{GT Type} & \textbf{Notes} \\
\midrule
TartanAir~\cite{tartanair2020iros} \textcolor{red}{} & Synthetic (indoor/outdoor) & 306K & $640\times480$ &  Simulated & Comprises 30 photo-realistic environments with diverse styles \\ 
FSD~\cite{wen2025stereo} \textcolor{red}{} & Synthetic (indoor/outdoor) & 1.04M & $1280\times720$ &  Simulated & Generated using NVIDIA Omniverse with diverse 3D assets \\
KenBurns~\cite{Niklaus_TOG_2019} \textcolor{red}{} & Synthetic (indoor/outdoor) & 76K & $512\times512$ &  Simulated & Synthetic dataset for view synthesis \\ 
Spring~\cite{Mehl2023_Spring} \textcolor{red}{} & Synthetic (indoor/outdoor) & 5K & variable &  Simulated & High-detail benchmark for stereo, optical flow, and scene flow \\
HyperSim~\cite{piccinelli2024unidepth} \textcolor{red}{} & Synthetic (indoor) & 77400 & $1024\times768$ &  Simulated & 461 indoor scenes with detailed per-pixel corresponding gt geometry \\
IRS~\cite{wang2021irslargenaturalisticindoor} \textcolor{red}{} & Synthetic (indoor) & 103K & $960\times540$ &  Simulated & Synthetic indoor dataset with diverse scenes and lighting conditions \\
Structured~\cite{yang2024depth} \textcolor{red}{} & Synthetic (indoor) & 76K & variable &  Simulated & Photo-realistic indoor scenes with comprehensive 3D structure annotations \\
Replica~\cite{ke2024repurposing} \textcolor{red}{} & Synthetic (indoor) & 18 scenes & variable(HDR) &  Simulated & High-fidelity 3D indoor scenes with HDR textures \\
UnrealStereo4K~\cite{li2024patchfusion} & Synthetic (indoor) & 1,600 & $3840\times2160$ &  Simulated & Photo-realistic indoor scenes for evaluating high-res depth models \\
Virtual KITTI 2~\cite{cabon2020virtual} & Synthetic (driving) & 21260 & $1242\times375$ &  Simulated & Synthetic clone of KITTI with dense depth and segmentation; various weather \\
MVS-Synth~\cite{DeepMVS} \textcolor{red}{} & Synthetic (outdoor/urban) & 12000 & variable &  Simulated & Photorealistic GTA V scenes with complete disparity maps \\
MatrixCity~\cite{li2023matrixcity} \textcolor{red}{} & Synthetic (outdoor) & 390K & $1920\times1080$ &  Simulated & Constructed using Unreal Engine 5's City Sample project \\
MidAir~\cite{Fonder2019MidAir} \textcolor{red}{} & Synthetic (outdoor) & 423K & $1024\times1024$ &  Simulated & Synthetic UAV dataset with diverse weather and terrain variations \\
GTA-SfM~\cite{wang2019flowmotiondepthnetworkmonocular} \textcolor{red}{} & Synthetic (outdoor) & 19K & variable (HD) & Simulated & Designed for SfM with large camera motions; diverse scenes and conditions \\
Sintel~\cite{Butler:ECCV:2012} & Synthetic (outdoor) & 1064 & $1024\times436$ & Simulated & Derived from 23 animated scenes of the open‑source 3D film “Sintel” \\ \hline
NYU Depth v2~\cite{Silberman2012IndoorSA} & Indoor & 1449 (dense) & $640\times480$ &  Kinect RGB-D & 464 scenes (home/office); densely sampled video frames also available \\
SUN RGB-D~\cite{song2015sun} & Indoor & 10,335 & $\approx 640\times480$ &  Kinect/RealSense & Merged from NYU, Berkeley B3DO, SUN3D, etc.; diversity of indoor scenes \\
ScanNet~\cite{dai2017scannet} & Indoor & 1513 Scenes & $\approx 640\times480$ &  Kinect~V1 & Captured with Kinect V1 and post processed with RGB-D SLAM system. \\
ScanNet++~\cite{yeshwanth2023scannethighfidelitydataset3d} & Indoor & 1000+ Scenes & variable & FARO + RGB-D & A large scale dataset with 1000+ 3D indoor scenes  \\
Hammer~\cite{jung2023importanceaccurategeometrydata} & Indoor & 13 scenes($\approx$16K) & $750\times500$ & 3D‑scanner + multi‑sensor & Multi-modal dataset of depth maps from multiple sensors. \\
iBims-1~\cite{koch2018evaluationcnnbasedsingleimagedepth} & Indoor & 100 & $640\times480$ & DSLR + laser scanner & RGB-D dataset, especially designed for testing single-image depth estimation. \\
Taskonomy~\cite{bhat2023zoedepth} \textcolor{red}{} & Indoor & 4.5M & $512\times512$ &  Aligned 3D meshes & The dataset consists of over 4.6 million images from 537 different buildings. \\
ARKitScenes~\cite{yin2023metric3d} \textcolor{red}{} & Indoor & 5047 Scans & $1920\times1440$ &  LiDAR + laser scanner &  Includes high-res and low-res depth maps \\
Make3D~\cite{fu2024geowizard} \textcolor{red}{} & Outdoor & 534 & $2272\times1704$ &  Laser scanner & Early dataset (2009) of outdoor scenes; sparsely sampled depth (posts) \\
DIODE~\cite{diode_dataset} & Indoor/outdoor & 27858 & $1024\times768$ &  FARO Focus 360 scanner & Indoor and outdoor dataset share the same sensor in 30 different scenes.\\
ETH3D~\cite{8954208}    & Indoor/outdoor & 96  & $752\times480$ & Synchronized RGB‑D & Includes synchronized global‑shutter color and depth, full trajectory GT. \\ \hline
KITTI~\cite{geiger2013vision} & Outdoor (driving) & 22k & $1242\times375$ &  LiDAR (sparse) & Driving scenes, various lighting; sparsity $\sim$5\% depth pixels per image \\
A2D2~\cite{geyer2020a2d2} & Outdoor (driving) & 196k & $1928\times1208$ &  LiDAR (sparse) & Driving scenes, using 6 cameras and 5 Velodyne VLP-16 LiDAR sensors \\
Argoverse2~\cite{li2024patchfusion} \textcolor{red}{} & Outdoor (driving) & 1000 sequences & $1550\times2048$ &  LiDAR (sparse) & From 7 ring cameras and 2 stereo cameras \\
Waymo~\cite{Sun_2020_CVPR} \textcolor{red}{} & Outdoor (driving) & 1950 sequences & $1920\times1280$ &  LiDAR (dense) & Includes data from 5 cameras and 5 LiDAR sensors per vehicle \\
DDAD~\cite{packnet} & Outdoor (driving) & 19680 & $1936\times1216$ & LiDAR (sparse) &  Contains monocular videos and accurate ground-truth depth \\
Nuscenes~\cite{nuscenes2019} & Outdoor (driving) & 1000 scenes & $1600\times900$ & LiDAR (sparse) & A public large-scale dataset for autonomous driving \\
MegaDepth~\cite{MegaDepthLi18}  & Landmark (outdoor) & 130k & variable (HD) &  SfM (multi-view) & Diverse landmarks, pseudo-depth (scale per scene), heavy occlusions \\ \hline
BlendedMVS(G)~\cite{yao2020blendedmvs} \textcolor{red}{} & Internet (wild) & 110K & variable & Multi-view stereo + recon & Large-scale dataset for solving multi-view geometry problems \\
DIW~\cite{fu2018deep} & Internet (wild) & 495K (sparse) & variable &  Human ordinal labels & Random internet photos with relative depth pairs annotated \\
ReDWeb~\cite{Xian_2018_CVPR}  & Internet (wild) & 3600 & variable &  Human ordinal + stereo & Diverse images with dense depth by stereo+manual clean-up (approx. metric) \\
Depth in the Wild~\cite{chen2020oasislargescaledatasetsingle} & Internet (wild) & 50K & variable &  SfM + manual & Large-scale dataset with pseudo depth, normals, boundaries for varied scenes \\
\bottomrule
\end{tabular}
}
\end{table*}

In the era of foundation models, combining datasets from diverse domains has become increasingly important. Foundation models~\cite{birkl2023midas,wang2025moge2accuratemonoculargeometry, wang2025moge,piccinelli2025unidepthv2universalmonocularmetric,yin2023metric3d, piccinelli2024unidepth, Hu_2024} typically leverage mixed-domain datasets to learn generalizable representations, which significantly enhancing robustness and adaptability across real-world scenarios. Several modern depth estimation frameworks also utilize hybrid and multi-stage training strategies~\cite{yang2024depth,depth_anything_v2,bochkovskii2024depth,bhat2023zoedepth}. These methods first undergo pretraining on extensive multi-domain datasets to build foundational capabilities, followed by fine-tuning on synthetic data, pseudo-labels generated by teacher models, or specialized domain-specific datasets with more stringent loss functions. Such a two-stage training paradigm markedly improves the qualitative detail results and generalizability of model.

\section{Depth Estimation Prior to Foundation Models} 
\label{sec:depth prior}
\begin{figure*}[t]
    \centering
    \includegraphics[width=0.99\linewidth]{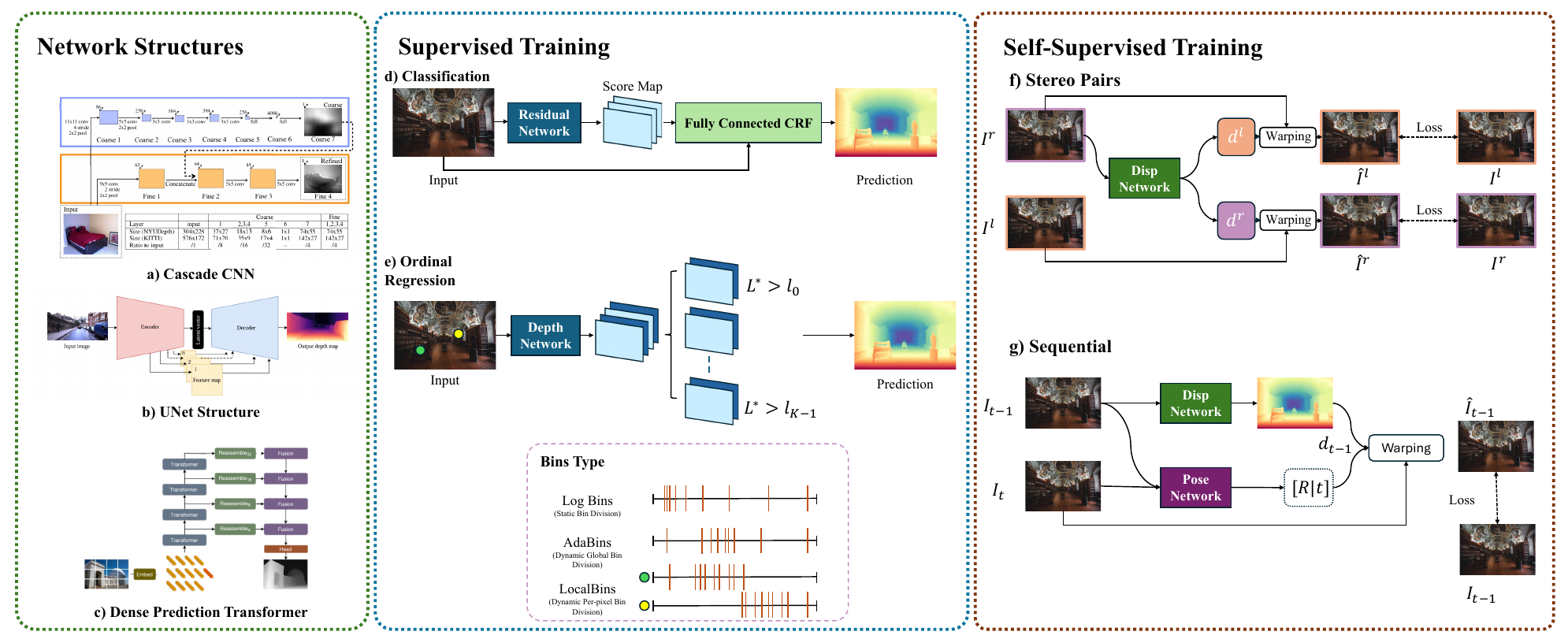}
    \caption{Overview of commonly used model structures (CNN, ResNet, ViT), training modes (supervised, unsupervised with and without pose annotations), and inference manners for depth estimation (regression, depth bin classification, and diffusion-based approaches).}
    \label{fig:before_lm}
\end{figure*}

In this section, we review the evolution of monocular depth estimation methods before the advent of large foundation models, as illustrated in Fig.~\ref{fig:before_lm}. We begin by discussing key advancements in depth estimation through \emph{supervised learning} (Sec.~\ref{sec: supervised}), covering both backbone architectures (Sec.~\ref{sec: structure}) and training objectives (Sec.~\ref{sec: definition}). Next, we introduce the \emph{self-supervised learning} paradigm (Sec.~\ref{sec: self-supervised}), which removes the reliance on ground-truth depth by leveraging view consistency losses. 

\subsection{Supervised Depth Estimation Methods}
\label{sec: supervised}
The supervised learning paradigm dominated the early development of monocular depth estimation. In this approach, a single RGB image is input into a carefully designed neural network, which predicts 3D representations such as depth maps, point clouds, or discrete depth bins. To guide the network in learning accurate depth predictions, various loss functions are employed, ranging from pixel-wise regression losses to geometric and structural consistency losses. 

\subsubsection{{Architectural Designs: From CNNs to Vision Transformers}}
\label{sec: structure}

In the early stages of monocular depth estimation, Eigen \textit{et al.}~\cite{eigen2014depth,eigen2015predicting} trained a \textit{multi-scale convolutional neural network (CNN)} on the NYU-v2~\cite{Silberman2012IndoorSA} and KITTI~\cite{geiger2013vision} datasets. To achieve dense and detailed depth prediction, their architecture consisted of two cascaded networks: a coarse network for predicting coarse depth structures and a fine network for refining local details. This approach set foundational benchmarks that subsequent methods built upon.

With the advent of ResNet~\cite{he2016deep}, Laina \textit{et al.}~\cite{laina2016deeperdepthpredictionfully} introduced a \textit{ResNet-50 encoder with up-projection blocks} for decoding dense depth, demonstrating that \textit{deeper networks improve performance and generalization}. This insight led many subsequent works~\cite{xian2018monocular,Lee_2019_CVPR,lienen2021monoculardepthestimationlistwise,lee2018single,MegaDepthLi18,abarghouei18monocular,xu2018structuredattentionguidedconvolutional,patil2022p3depthmonoculardepthestimation,you2021interpretabledeepnetworksmonocular,srinivasan2018aperturesupervisionmonoculardepth,gur2020singleimagedepthestimation}  to design progressively deeper architectures based on the ResNet backbone for depth estimation. 

However, these architectures~\cite{chen2019attentionbasedcontextaggregationnetwork, miangoleh2021boostingmonoculardepthestimation}, which were based on convolutional neural networks, struggled to balance global context with fine-grained details. Due to their limited receptive fields, convolutional networks often failed to capture global context and large-scale scene structures, resulting in artifacts such as ``banding'' or inconsistent depth across distant regions.

The application of \textit{Vision Transformers} (ViT) in this domain has significantly improved these limitations. Ranftl \textit{et al.}~\cite{ranftl2021vision} proposed the \textit{Dense Prediction Transformer (DPT)}, which utilizes a Vision Transformer encoder (pre-trained on ImageNet) to capture global context, while employing a convolutional decoder with multi-scale features from the encoder to recover fine-grained details. This approach achieved substantial improvements, with up to a 28\% reduction in relative error compared to CNN baselines across multiple benchmarks, and generated more spatially coherent depth maps. Since then, recent methods~\cite{ning2023trap, li2023patchfusionendtoendtilebasedframework, guizilini2023zeroshotscaleawaremonoculardepth, guo2025depthcamerazeroshotmetric, kong2023robodepthrobustoutofdistributiondepth, Papa_2023, DBLP:journals/sensors/IbrahemSK22c,yang2024depth,yang2024depthv2} have adopted hybrid ViT encoder and CNN decoder designs to effectively combine global and local information for dense depth estimation.

\subsubsection{Training Objectives}
\label{sec: definition}
Depth estimation has traditionally been framed as a discriminative learning problem, with several learning objectives explored: regression, classification, and ordinal regression.  The predicted depth map  $D \in \mathbb{R}^{H \times W}$ is generated from an input image $I$ using a trainable neural network $f_{\theta}$, as follows:  

\begin{equation}
D = f_{\theta}(I), 
\end{equation}
In the following, $D^*$ denotes the ground-truth depth. 

\paragraph{Regression} Depth estimation can naturally be formulated as a regression task. Standard losses include the $\mathcal{L}_1$ or $\mathcal{L}_2$ norms: 

\begin{align}
\mathcal{L}_{\text{1}}(D, D^*) &= \frac{1}{n} \sum_{i=1}^{n} \bigl|\,y_i - y^*_i\bigr|, \\
\mathcal{L}_{\text{2}}(D, D^*) &= \frac{1}{n} \sum_{i=1}^{n} (y_i - y^*_i)^2,
\end{align}
where $y_i$ and $y^*_i$ denote the predicted and ground-truth depth values at pixel $i$, and $n$ is the number of valid pixels. Although conceptually simple, these losses often lead to blurred depth discontinuities and fail to preserve fine structural details in the scene.

To balance small and large residuals (losses),  Laina \textit{et al.}~\cite{laina2016deeperdepthpredictionfully} proposed the inverse Huber loss (or \textit{berHu loss}):
\begin{equation}
\mathcal{B}(x) =
\begin{cases}
\lvert x\rvert, & \lvert x\rvert \le c,\\[6pt]
\dfrac{x^{2} + c^{2}}{2\,c}, & \lvert x\rvert > c,
\end{cases}
\label{eq:berhu}
\end{equation}
where $x = y_i - y^*_i$ is the residual, and the threshold $c$ is empirically set as $c = \tfrac{1}{5} \max_i \lvert y_i - y^*_i \rvert$. This loss behaves like $\mathcal{L}_1$ loss for low error pixels and switches to $\mathcal{L}_2$ for larger discrepancies. This hybrid formulation improves both convergence and boundary accuracy. 
Later works such as CLIFFNet~\cite{wang2020cliffnet} introduced hierarchical losses leveraging intermediate features to enhance multi-scale consistency beyond pixel-level accuracy.

\paragraph{Scale-invariant Loss} 
Monocular depth estimation is inherently ill-posed, as different 3D scenes of varying scale can produce identical 2D images. To mitigate this ambiguity, Eigen \textit{et al.}~\cite{eigen2014depth} proposed a scale-invariant loss: 
\begin{equation}
\mathcal{L}_{\text{si}}(D, D^*) = \frac{1}{n} \sum_{i=1}^n d_i^2 - \frac{\lambda}{n^2} \left( \sum_{i=1}^n d_i \right)^2,
\end{equation}
where $d_i = \ln y_i - \ln y_i^*$, and $\lambda \in [0,1]$ is a hyperparameter.
When $\lambda =1$, the loss becomes fully scale-invariant, encouraging accurate relative depth rather than absolute depth magnitude.

\paragraph{Affine-invariant Loss} Building on scale-invariant loss, recent methods~\cite{yin2020diversedepthaffineinvariantdepthprediction,garg2019learningsinglecameradepth,wang2025moge,yang2024depth,depth_anything_v2,he2024lotus,ke2024repurposing} adopt an \textit{affine-invariant formulation} that aligns predicted depth with ground-truth via a global scale $s$ and translation $t$:
\begin{equation}
\mathcal{L}_{\mathrm{ssi}}(D,D^*) \;=\; \min_{s,t}\;\frac{1}{n}\sum_{i=1}^n \left\|s\,y_i + t - y^*_i\right\|.
\label{eq:affine-invariant}
\end{equation}
Parameters $s$ and $t$ are typically solved using least-squares. 
The aligned predictions are then evaluated using $\mathcal{L}_1$ $\mathcal{L}_2$ losses.

\paragraph{Classification}
Direct regression of continuous depth values is challenging. Inspired by human perception, which favors relative depth cues over exact values, Cao \textit{et al.}~\cite{cao2017estimating} reformulated the task as a classification problem. Depth values are discretized into predefined bins, and the network is trained to classify each pixel into a depth interval.

\paragraph{Ordinal Regression} 
Standard classification fails to capture the ordinal nature of depth. DORN~\cite{fu2018deep} introduced ordinal regression by dividing the depth range into
$K$ intervals between a near plane $\alpha$ and far plane $\beta$, using either uniform discretization (UD) or space-increasing discretization (SID):
\begin{align}
\text{UD: } & t_i = \alpha + (\beta - \alpha) \cdot \frac{i}{K}, \\
\text{SID: } & t_i = \exp\left[\log(\alpha) + \frac{\log(\beta / \alpha) \cdot i}{K}\right],
\end{align}
where $t_i$ defines the bin thresholds.

The model outputs an ordinal depth tensor $Y \in \mathbb{R}^{H \times W \times 2K}$, from which depth is recovered as:
\begin{align}
\hat{l}{(w, h)} &= \sum_{k=0}^{K-1} \mathbb{I}\left(\mathcal{P}^k_{(w, h)} \geq 0.5\right), \\
\hat{d}{(w, h)} &= \frac{t{\hat{l}{(w, h)}} + t{\hat{l}{(w, h)} + 1}}{2} - \xi, \\
\mathcal{P}^k{(w, h)} &= \frac{e^{y_{(w, h, 2k+1)}}}{e^{y_{(w, h, 2k)}} + e^{y_{(w, h, 2k+1)}}},
\end{align}
where $\xi$ ensures $\alpha + \xi = 1$, and $\mathbb{I}(\cdot)$ is the indicator function.
Furthermore, they also propose the ordinal regression loss which is defined as:
\begin{equation}
\mathcal{L}_{\text{ord}} = \frac{1}{n} \sum_{i=1}^{n} \left( \sum_{k=0}^{l_i - 1} \log(\mathcal{P}_i^k) + \sum_{k=l_i}^{K - 1} \log(1 - \mathcal{P}_i^k) \right).
\end{equation}

Building on this insight, and motivated by the observation that depth distributions can vary significantly across images-- ranging from narrow bands in tabletop scenes to broad ranges in corridor views-- a transformer head was introduced to \emph{jointly} predict a set of image-specific depth bin centers \(c(\mathbf b)=\{c_1,\dots,c_{B}\}\). To further capture pixel-level variations in depth range, LocalBins~\cite{bhat2022localbins} proposed predicting depth bins locally for each pixel, thereby achieving finer spatial granularity. Since then, several methods~\cite{wang2025scalable,shao2023iebins,lee2023slabins,li2024binsformer,zhu2023ha,she2024evitbins} have introduced variants of bin prediction strategies to further enhance performance.

\paragraph{Geometric Consistency Loss} 
Many approaches~\cite{qi2018geonet,qi2020geonet++,shao2023nddepth,piccinelli2023idisc,wofk2019fastdepth,yang2023gedepth,yuan2022neural,huynh2020guiding,ramamonjisoa2019sharpnet,wang2025moge,wang2025moge2accuratemonoculargeometry} have leveraged the geometric relationship between depth and surface normals to improve prediction accuracy by jointly estimating and enforcing consistency between the two. Depth and surface normals are intrinsically linked: the surface normal at a point is determined by the tangent plane of the local 3D surface, which can be inferred from the depth map; conversely, depth is constrained by the orientation of this tangent plane defined by the surface normal. By incorporating modules that estimate normals from depth and vice versa, consistency losses are introduced to align the predicted (or the groun-truth) and derived ones. A typical formulation of the consistency-enforced loss functions is:

\begin{align}
\mathcal{L}_{\text{depth}} &= \frac{1}{n} \sum_{i=1}^n \left( \| y_i - y_i^* \|_2^2 + \eta \cdot \| \hat{y}_i - y_i^* \|_2^2 \right), \\
\mathcal{L}_{\text{normal}} &= \frac{1}{n} \sum_{i=1}^n \left( \| \mathbf{n}_i - \mathbf{n}_i^* \|_2^2 + \lambda \cdot \| \hat{\mathbf{n}}_i - \mathbf{n}_i^* \|_2^2 \right),
\end{align}
where \(y_i\) and \(\hat{y}_i\) denote the initial and refined  depth predictions at pixel \(i\), and \(y_i^*\) is the corresponding ground-truth depth. Similarly, \(\mathbf{n}_i\) and \(\hat{\mathbf{n}}_i\) are the initial and refined predicted surface normals, while \(\mathbf{n}_i^*\) denotes the ground-truth normal. The weights \(\eta\) and \(\lambda\) control the contribution of the refined predictions.

\subsection{Self-Supervised Learning}
\label{sec: self-supervised}
Supervised training schemes for monocular depth estimation rely heavily on accurate ground-truth depth or point cloud annotations, which are often expensive and difficult to obtain. In contrast, humans can intuitively infer 3D scene structure and ego-motion without explicit geometric supervision. Motivated by this observation, a substantial body of work~\cite{yan2025synthetic, wu2025geodepth, moon2024ground, nguyen2024mining, wang2023planedepth, zhang2023lite, si2023fully, petrovai2022exploiting, gonzalez2021plade, guizilini20203d, poggi2020uncertainty, xian2018monocular, zhao2023gasmono, rodriguez2023lightdepth, zhang2023robust, han2023self, saunders2023self, varghese2023self, ji2021monoindoor, wang2021regularizing, liu2021self, li2021structdepth, chen2019self, bello2021self} has emerged around \emph{self-supervised} depth estimation. These methods eliminate the need for direct depth labels by leveraging pixel-level consistency losses as supervision signals. 

\subsubsection{Stereo Supervision}
Garg~\etal~\cite{garg2016unsupervisedcnnsingleview} pioneered unsupervised depth estimation using stereo image pairs, aiming to sidestep the need for labeled ground-truth depth. Given a rectified  stereo pair ${I_l, I_r}$, their network predicts the disparity $d_l$ of the left image $I_l$, which is then used to synthesize the right image $\hat{I}_r^l$ via image warping. The photometric reconstruction loss is used to supervise the network: 
\begin{equation}
\label{equ: photometric}
\mathcal{L}_\text{photo} = | I_r - \hat{I}_r^l |_2.
\end{equation}

Monodepth~\cite{godard2017unsupervisedmonoculardepthestimation} improved this framework by predicting both left-to-right and right-to-left disparities 
 ($d_l$, $d_r$),  enforcing left-right consistency, and introducing additional appearance and smoothness losses: 
\begin{align}
\mathcal{L}_{a} &= \frac{1}{N}\sum_{i,j}\left[\alpha\frac{1 - \text{SSIM}(\hat{I}_r^l, I_r)}{2} + (1 - \alpha) | \hat{I}_r^l - I_r |_1\right], \\
\mathcal{L}_\text{sm} &= \frac{1}{N} \sum_{i,j} \left(
\left| \partial_x d_l \right| e^{- \left| \partial_x I_l \right|} +
\left| \partial_y d_l \right| e^{- \left| \partial_y I_l \right|}
\right).
\end{align}

\subsubsection{Video Sequences}
While stereo-based self-supervision eliminates the need for depth labels, it still requires stereo camera setups, limiting scalability. To overcome this, Zhou~\etal~\cite{zhou2017unsupervised} introduced \emph{SfMLearner}, which leverages monocular video sequences and employs a second network to estimate relative camera poses. Their framework jointly learns depth and ego-motion from unlabeled videos via view synthesis. 

Given a sequence $S = \{I_0, I_1, ..., I_N\}$, one frame is designated as the target view $I_t$ and the remaining frames serve as source views $I_s$ ($s \neq t$).
The depth network $f_\theta$ predicts the disparity $\hat{D}_t$ for $I_t$, while the pose network $g_\phi(I_t, I_s)$ estimates the $4\times 4$ transformation matrix $T_{t \rightarrow s}$.
The target pixel $p_t$  can then be projected into the source frame as:
\begin{equation}
p_s \sim K T_{t \rightarrow s} \hat{D}_t(p_t) K^{-1} p_t,
\end{equation}
where $K$ is the camera intrinsic matrix. Photometric consistency loss (Eq.~\eqref{equ: photometric}) between $I_t$ and its warped reconstruction $\hat{I}_t$ are used to jointly supervise depth and pose estimation. 

Monodepth2~\cite{monodepth2} further refined this pipeline by introducing an auto-masking strategy to handle dynamic objects and occlusions, and by adopting a ResNet backbone pretrained on ImageNet. Monodepth2 demonstrated that with deeper architectures and strong initialization, self-supervised methods can match or even outperform some supervised models on standard benchmarks. This paradigm has been widely adopted, with continued innovations in architecture design~\cite{lyu2021hr,he2022ra,chen2019self,patil2020don,dai2020self,cheng2020s,zhao2022monovit}, uncertainty modeling~\cite{poggi2020uncertainty,watson2019self}, dynamic scene understanding~\cite{sun2023sc,shao2022self}, and integration with downstream applications~\cite{zhou2021self,johnston2020self,madhuanand2021self,zhang2023lite}.

\subsection{Summary}
Prior to the emergence of large foundation models, research in monocular depth estimation had largely converged in both architectural design and training objectives. Architecturally, hybrid CNN-ViT models-- particularly the DPT-style architecture-- demonstrated superior performance and became a foundational backbone for several subsequent foundation models~\cite{birkl2023midas, depth_anything_v2,yang2024depth,piccinelli2024unidepth,piccinelli2025unidepthv2universalmonocularmetric,bochkovskii2024depth,guizilini2023zeroshotscaleawaremonoculardepth,yin2023metric3d,Hu_2024,xue2026depthart,liu2026foundationgeo}.
On the training side, scale-invariant loss functions gained widespread adoption for relative depth estimation and have been carried over into the training regimes of foundation models~\cite{ke2024repurposing,wang2025moge,yang2024depth,depth_anything_v2,fu2024geowizard,yin2021learning,he2024lotus,garcia2025fine,xu2024matters} as well. In addition, surface normal regularization has become a common strategy for improving 3D reconstruction fidelity, and this practice has similarly influenced the training of modern foundation models~\cite{wang2025moge, wang2025moge2accuratemonoculargeometry,garcia2025fine,xu2024matters,Hu_2024,zhao2024metric}.

\section{Depth Foundation Model}
\label{sec:Depth Foundation Model}

The remarkable success of large foundation models in natural language processing and image/video understanding has recently catalyzed a paradigm shift in monocular depth estimation, leading to the emergence of \textit{Depth Foundation Models}. These models harness the power of large-scale, pre-trained foundation models (e.g., DINOv2~\cite{oquab2023dinov2}, DINOv3~\cite{simeoni2025dinov3}, Stable Diffusion~\cite{rombach2022high}) and are further fine-tuned on extensive RGB-D datasets. Their primary goal is to achieve accurate depth estimation on in-the-wild images. Typically, Depth Foundation Models are designed with the following objectives:

\begin{enumerate}
\item \textbf{Zero-shot robustness.} A single model should generalize effectively to previously unseen inputs, delivering reliable depth estimations across a wide variety of camera types (e.g., pinhole, wide-angle, fisheye) and scenes (e.g., indoor, outdoor, in-the-wild, object-centric, and stylized environments), without requiring additional fine-tuning.
\item \textbf{Detail preservation.} The predicted depth maps should be both highly accurate and capable of preserving fine-grained geometric details, characterized by sharp, structurally complete representations.
\end{enumerate}
In this section, we begin by summarizing recent advancements focused on unifying and integrating diverse datasets across existing approaches for training depth foundation models. We then categorize the latest methods into \emph{Discriminative Models} and \emph{Generative Models} based on differences in their prediction paradigms, offering a comprehensive review of the progress achieved in each category.

\begin{figure*}[htbp]
    \centering
    \includegraphics[width=1.0\linewidth]{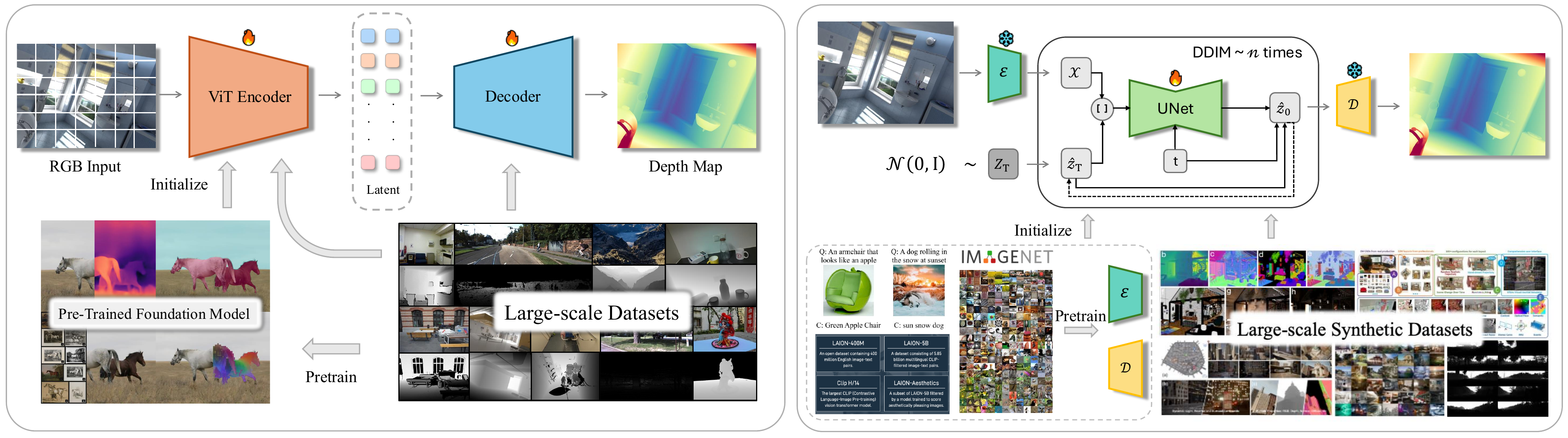}
    \caption{Illustration of discriminative (left) and generative (right) frameworks for monocular depth estimation. Discriminative methods typically leverage a Vision Transformer~\cite{dosovitskiy2021imageworth16x16words} (ViT) encoder initialized with a pre-trained foundation model (e.g., DINOv2~\cite{oquab2023dinov2}) and trained on large-scale diverse datasets to directly regress depth maps. In contrast, generative methods utilize VAE~\cite{kingma2022autoencodingvariationalbayes} structure, iteratively refining depth predictions through a denoising process conditioned on synthetic datasets, thereby effectively capturing complex details and structural coherence.}
    \label{fig:foundation model structure}
\end{figure*}

\subsection{Dataset Collection and Unification}

\paragraph{Mixed Data Alignment} RGB-D datasets often exhibit considerable heterogeneity in their depth annotations, which may include absolute depth (e.g., from laser-based sensors or stereo cameras with known calibration), depth with an unknown scale (e.g., from Structure-from-Motion), or disparity maps (e.g., from stereo setups without calibration). To address this challenge, MiDaS~\cite{ranftl2020towards} introduced the affine-invariant loss (Eq.~\eqref{eq:affine-invariant}), which is robust to scale and shift variations, and tackles the primary sources of incompatibility among different datasets. This approach enables the model to effectively integrate diverse data sources while utilizing the full spectrum of available information.

\paragraph{Emergence of High-Quality Synthetic Data}
Depth labels acquired in real-world scenes using laser-based sensors (RGB-D cameras or LiDAR) are often affected by noise, particularly in transparent or highly reflective regions, and may have missing values. In recent years, the creation of photorealistic rendering datasets with accompanying depth information has surged, utilizing simulators or game engines such as Unreal Engine, Unity, and Isaac Sim~\cite{mittal2023orbit}. These synthetic datasets, thanks to their photorealism and accurately annotated depth labels, significantly enhance the robustness of depth estimation models. 

\paragraph{Data in Challenging Conditions}
Several recent efforts have focused on depth estimation under challenging or out-of-distribution (OOD) conditions, such as underwater environments, adverse weather (e.g., fog, rain), or scenes dominated by non-Lambertian surfaces (e.g., reflective or translucent materials). In these scenarios, obtaining ground-truth depth annotations is particularly difficult or even impractical. To mitigate this, researchers~\cite{zhang2024atlantis, shim2024sediff, tosi2024diffusion} have turned to cutting-edge text-to-image diffusion models with depth-aware control mechanisms, enabling the generation of high-quality synthetic images that are aligned with specific depth structures. These models maintain geometric consistency while synthesizing realistic visual content tailored to rare or complex environments.

\subsection{Discriminative Methods}

Many existing depth foundation models formulate this task as a discriminative learning problem, directly learning a mapping from images to depth values. In this area, before the emergence of Vision Transformers (ViTs), early zero-shot monocular depth estimation methods-- such as MegaDepth~\cite{MegaDepthLi18}, MiDaS~\cite{ranftl2020towards}, and LRSI~\cite{yin2021learning}-- primarily relied on convolutional neural networks (CNNs) as backbones. These networks were typically pre-trained on large-scale image classification datasets (e.g., ResNet-101, DenseNet-161) to enhance feature extraction. While these CNN-based methods demonstrated promising performance on unseen datasets, their depth estimation accuracy remained limited. In particular, they struggled with mixed-dataset training and often failed to outperform models trained on domain-specific data, showing poor scalability as the data size increased.

To overcome these limitations, MiDaS v3.1~\cite{birkl2023midas} introduced the first depth foundation models using the ViT-based DPT architecture, which achieved superior performance, particularly due to its ability to capture global context and long-range spatial relationships. Subsequently, the ViT-based architecture became the new standard and was widely adopted in foundational depth models~\cite{yang2024depth, depth_anything_v2, piccinelli2024unidepth,piccinelli2025unidepthv2universalmonocularmetric, wang2025moge,wang2025moge2accuratemonoculargeometry,liu2024sm4depth}. A representative discriminative model structure is shown in Fig.~\ref{fig:foundation model structure}.

\subsubsection{\textbf{Relative Depth Estimation}}

\paragraph{Transformer Architectures} Since the introduction of DPT~\cite{ranftl2021vision}, which utilizes transformers for dense predictions, it has been widely adopted for building foundation models in relative depth estimation. Building on the DPT architecture, the MiDaS series (v2~\cite{Ranftl2019}, v3~\cite{ranftl2020towards}) were the first to train the model on 1.1 million images from diverse sources (including ReDWeb~\cite{xian2018monocular}, MegaDepth~\cite{MegaDepthLi18}, WSVD~\cite{wang2019web}, and others), producing relative depth estimations that effectively captured geometry across a wide range of inputs. These models demonstrated that a single model could generalize well to indoor, outdoor, and even artistic images, establishing DPT as a foundational model architecture and kickstarting the development of depth foundation models.

\paragraph{Scale-up Learning} {Depth Anything} \cite{yang2024depth} uses the DPT architecture and embodies a foundation model that focuses not on novel architectures but on a massive ``data engine'' to enable in-the-wild generalization. It collected 1.5 million labeled and 62 million unlabeled images, annotated with pseudo-depth from model ensembles. An important factor contributing to its success is the use of pre-trained visual representations from the vision foundation model DINOv2. This allowed the model to achieve exceptional zero-shot relative-depth performance across six public datasets and diverse real-world images. However, the Depth Anything model still struggles to deliver sharp results. To address this, {Depth Anything V2} \cite{depth_anything_v2} replaced all real-world labels with synthetic ones to train a teacher model, using the pre-trained DINOv2 model~\cite{oquab2023dinov2} to initialize the ViT encoder's weights.

\paragraph{Point-based Representation} Rather than directly predicting relative depth, Wang \textit{et al.} proposed {MoGe}~\cite{wang2025moge}, a foundation model built with a DPT architecture that predicts a per-pixel \emph{affine-invariant 3D pointmap} instead of depth. This approach bypasses the need for camera intrinsics, allowing for high-quality point reconstruction. MoGe incorporates dedicated training objectives, including affine-invariant global point supervision similar as Eq.~\eqref{eq:affine-invariant} but applied in point space, multi-scale local geometry loss, and normal loss, all of which improve surface reconstruction and local details. The model uses a DPT-style architecture with a ViT encoder initialized with DINOv2~\cite{oquab2023dinov2}, and it was trained on 9 million images spanning indoor, outdoor, aerial, and synthetic domains. This training yields a highly transferable geometry prior that can be quickly calibrated to metric scale or applied to downstream 3D vision tasks.

\begin{figure}[!htb]
    \centering
    \includegraphics[width=1.0\linewidth]{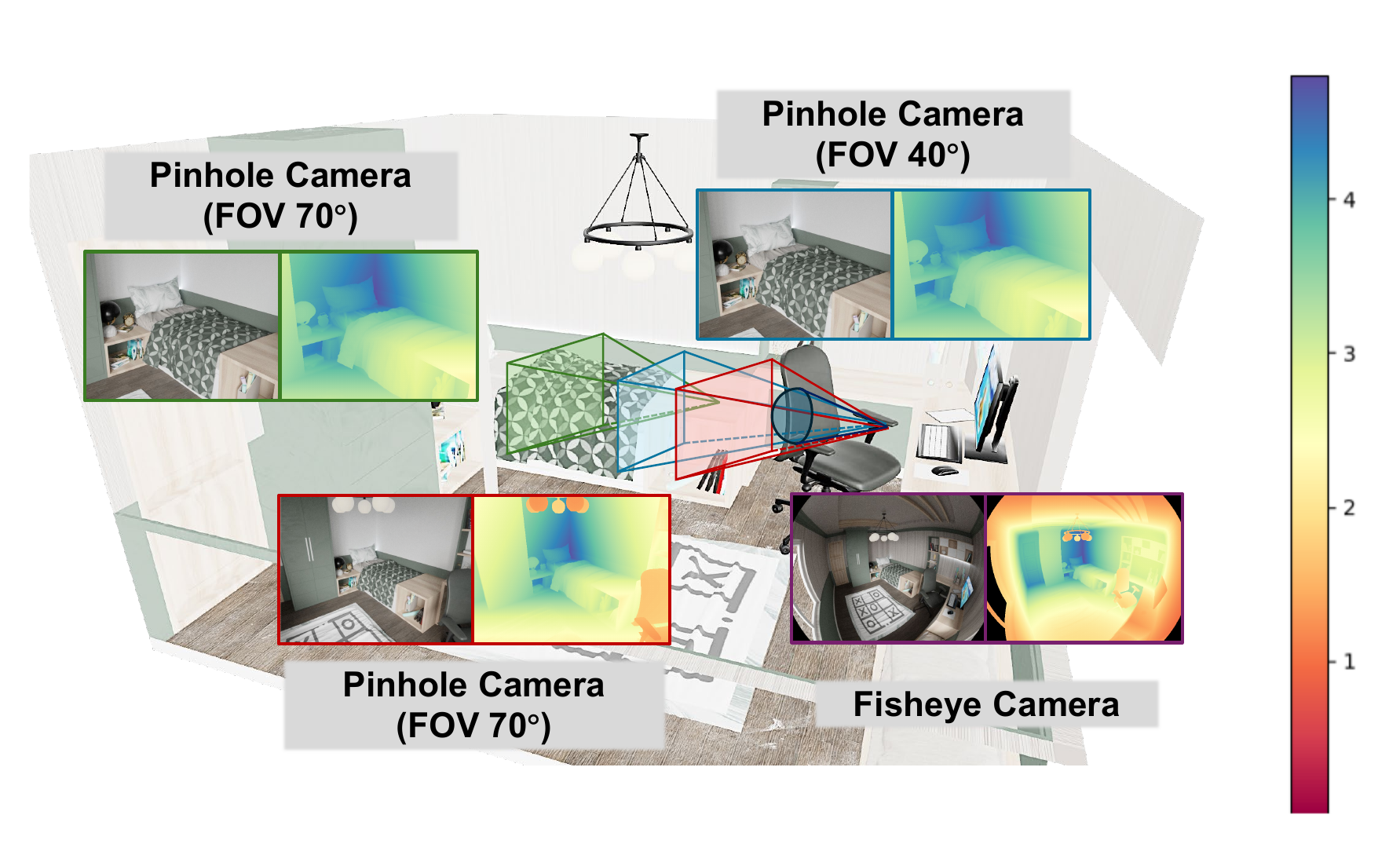}
    \caption{
\textbf{Illustration of metric depth ambiguity across different camera models and intrinsics.} We render the same indoor scene using four different cameras: three pinhole cameras (\textcolor[rgb]{0.3,0.5,0.2}{green}, \textcolor[rgb]{0.0,0.2,0.7}{blue}, \textcolor[rgb]{0.75, 0, 0}{red}) with varying fields-of-view (FoV), and one fisheye camera (\textcolor[rgb]{0.5,0.0,0.5}{purple}). 
Three cameras are placed at the same position (\textcolor[rgb]{0.0,0.2,0.7}{blue}, \textcolor[rgb]{0.75, 0, 0}{red}, \textcolor[rgb]{0.5,0.0,0.5}{purple}), yet they produce significantly different appearance patterns due to variations in camera intrinsics and distortion, despite sharing the same underlying geometry.
In contrast, pinhole cameras with different FoVs (\textcolor[rgb]{0.3,0.5,0.2}{green} vs. \textcolor[rgb]{0.0,0.2,0.7}{blue}) produce visually similar images but correspond to different depth distributions.  
}
\label{fig:metric_ambiguity}
\end{figure}

\vspace{0.1in}\subsubsection{\textbf{Metric Depth Estimation}}
\paragraph{Challenges} In addition to foundation models for relative depth estimation, many research efforts~\cite{wang2025moge2accuratemonoculargeometry,piccinelli2024unidepth,piccinelli2025unidepthv2universalmonocularmetric,bochkovskii2024depth,yin2023metric3d,Hu_2024} have sought to tackle the more challenging task of estimating metric depth from a single image. This problem is fundamentally more difficult due to inherent ambiguities stemming from unknown camera intrinsics (e.g., focal length, sensor type) and substantial domain shifts across diverse scenarios (e.g., varying depth ranges in indoor, outdoor, object-centric, real, or synthetic environments).

As illustrated in Fig.~\ref{fig:metric_ambiguity}, we render the same indoor scene using four different cameras. The setup includes two camera models-- pinhole (\textcolor[rgb]{0.3,0.5,0.2}{green}, \textcolor[rgb]{0.0,0.2,0.7}{blue}, \textcolor[rgb]{0.75, 0, 0}{red}) and fisheye (\textcolor[rgb]{0.5,0.0,0.5}{purple})-- with the pinhole cameras further configured using two different fields of view (FoV).

When placed at the same physical location (\textcolor[rgb]{0.0,0.2,0.7}{blue}, \textcolor[rgb]{0.75, 0, 0}{red}, \textcolor[rgb]{0.5,0.0,0.5}{purple}), these cameras-- with different camera models or FoV settings-- produce drastically different visual appearances, despite observing the same scene geometry and sharing identical depth values for the same objects. Conversely, two pinhole cameras with different FoVs (\textcolor[rgb]{0.3,0.5,0.2}{green} and \textcolor[rgb]{0.0,0.2,0.7}{blue}) and placed at different positions can yield visually similar images, even though they correspond to different underlying metric depths.

These observations underscore the intrinsic difficulty of monocular metric depth estimation: models must learn to generalize across diverse camera intrinsics and distortion types, while maintaining accurate and consistent geometric understanding.

\paragraph{Modeling Camera Intrinsics} A prominent line of research focuses on incorporating camera models into the architectural design to reduce the ambiguity introduced by varying camera intrinsics. Metric3D~\cite{yin2023metric3d} introduced a shared \emph{canonical camera space transformation} module that mitigates metric ambiguities by normalizing input images into a unified camera intrinsic system. Trained on large-scale datasets from diverse sources, it demonstrates strong zero-shot generalization.
Building on this, Hu \emph{et al.} proposed Metric3D v2~\cite{Hu_2024}, which incorporates joint constraints from both depth and surface normals, further improving zero-shot performance. However, these methods rely on known intrinsic parameters to perform the canonical transformation.

To overcome this limitation, UniDepth~\cite{piccinelli2024unidepth} introduces a self-prompted camera module that infers a dense representation of camera intrinsics, along with a geometric-invariance loss that helps disentangle camera-specific effects from depth predictions. Its successor, UniDepthV2~\cite{piccinelli2025unidepthv2universalmonocularmetric}, simplifies the architecture by replacing the spherical harmonic tokenization with a more efficient design based on a DINOv2-initialized Vision Transformer and sine positional encoding-- yielding improved performance with reduced complexity.

Similarly, DepthPro~\cite{bochkovskii2024depth} incorporates a learned focal-length prediction head, enabling metric depth estimation without access to explicit camera metadata. It further leverages synthetic data to enhance performance and preserve fine details.

These strategies-- ranging from canonical camera transformations to self-prompted camera intrinsics and learned focal-length estimation-- enable robust and universal metric depth estimation across diverse imaging conditions.

\paragraph{Repurposing Relative Depth Estimation for Metric Depth Prediction}
Another line of research aims to adapt relative depth estimation models for predicting metric depth. A representative example is ZoeDepth~\cite{bhat2023zoedepth}, which builds upon a pre-trained relative depth model by introducing a metric bin module that adaptively predicts bin centers to account for image variability. Once fine-tuned on NYU and KITTI datasets, the model demonstrates strong generalization across diverse scenes.

More recently, MoGe-2~\cite{wang2025moge2accuratemonoculargeometry} extends its predecessor MoGe~\cite{wang2025moge} by decoupling affine-invariant relative depth estimation and global metric scale estimation within a unified framework. It employs a lightweight MLP branch to predict a global scale factor for accurate metric depth. In addition, MoGe-2 introduces a data refinement pipeline that filters noisy real-world depth data using high-quality synthetic labels, leading to sharper object boundaries and improved depth detail. These refined data have been shown to significantly enhance depth estimation quality.

Building on relative-to-metric transfer, FoundationGeo~\cite{liu2026foundationgeo} reveals that metric errors arise not only from global scale ambiguity, but also from spatially varying scale drift and residual ray-direction bias. It further identifies focal-length coverage mismatch as a key bottleneck for zero-shot generalization. Accordingly, FoundationGeo introduces pixel-wise scale and ray-direction correction fields, together with targeted multi-focal synthetic data. Combined with its two-stage training strategy and broad multi-domain data coverage, FoundationGeo further raises the model’s performance ceiling while exhibiting stable generalization across domains and camera models.

\subsection{Generative Methods}

Originally developed for probabilistic generative modeling, diffusion models have recently gained traction in depth estimation due to their capacity to capture complex data distributions and preserve fine details. An increasing number of works~\cite{ke2024repurposing, fu2024geowizard, xu2024matters, garcia2025fine, zhang2024betterdepth, he2024lotus, song2026depthmaster,xupixel} leverage pre-trained diffusion models to enhance the realism and reliability of depth predictions. As shown in Fig.~\ref{fig:foundation model structure} (b), Marigold~\cite{ke2024repurposing} is a representative example, fine-tuned from Stable Diffusion for depth estimation. It concatenates RGB image latents with noise along the channel dimension, enabling the U-Net to iteratively denoise the input noise and produce a refined depth map. Notably, despite being trained on limited synthetic data, Marigold demonstrates competitive performance, particularly in fine-grained details.

\paragraph{Efficiency and Detail Enhancement}
Due to the inherent multi-step denoising process and test-time ensembling used to address sampling uncertainty, Marigold suffers from long inference times: approximately 24 seconds for a 578×578 input. To improve efficiency, recent methods reformulate depth estimation as a single-step diffusion process. GenPercept~\cite{xu2024matters} and Diffusion-E2E~\cite{garcia2025fine} investigate the role of the denoising scheduler and propose deterministic one-step alternatives. By reframing depth estimation as a direct prediction task, these approaches achieve faster inference while allowing the integration of pixel-level supervision via affine-invariant loss functions.

While diffusion models are adept at generating detailed outputs, fine structures may still degrade during task-specific adaptation-- potentially due to catastrophic forgetting~\cite{he2024lotus}. To address this, Lotus~\cite{he2024lotus} introduces a task switcher that alternates between generating depth/normal predictions and reconstructing input images, encouraging the preservation of structural and textural cues. Additionally, GeoWizard~\cite{fu2024geowizard} exploits the flexibility of diffusion models to jointly estimate depth and surface normals, enabling mutual information sharing and improved consistency between modalities. 

\paragraph{Beyond Diffusion}
Beyond conventional diffusion models, a few works draw inspiration from visual autoregressive modeling (VAR)~\cite{tian2024visual}, reformulating depth estimation as a sequence of next-scale predictions across multiple resolutions\cite{wang2025scalable}. However, their robustness and performance ceilings remain under early investigation.

\paragraph{Discussions}
Generative models typically yield sharper and more structurally coherent outputs, whereas discriminative models offer superior computational efficiency. Motivated by their complementary strengths, several hybrid strategies have been proposed. For instance, BetterDepth~\cite{zhang2024betterdepth} uses diffusion models as plug-and-play refiners to enhance coarse outputs from pretrained discriminative networks (e.g., DPT~\cite{ranftl2021vision}), improving fine-detail quality.

Furthermore, GenPercept and Diffusion-E2E~\cite{garcia2025fine} show that reducing diffusion to a single-step sampling scheme—while employing discriminative-style loss functions—boosts performance. These findings reflect a growing trend of repurposing diffusion models not as generative samplers but as deterministic predictors or feature extractors, thus bridging the gap between generative flexibility and discriminative efficiency.

\begin{table*}[ht]
  \vspace{-3.5mm}
  \caption{Unified evaluation of different methods. This benchmark is majorly for zero-shot evaluation. The best results are highlight in \textbf{bold}, and the second-best ones are \underline{underlined}. \textcolor{gray!50}{Gray numbers} denote models trained on respective benchmarks and thus excluded from ranking.}
  \label{table:benchmark}
  \centering
  \renewcommand{\arraystretch}{1.25}
  \adjustbox{width={\linewidth},keepaspectratio}{
    \begin{tabular}{l|cc|cc|cc|cc|cc|cc|cc|cc|cc|cc}
    \bottomrule
        \multirow{2}{*}{\textbf{Depth Align}} & \multicolumn{2}{c}{NYUv2} & \multicolumn{2}{c}{KITTI} & \multicolumn{2}{c}{ETH3D} & \multicolumn{2}{c}{iBims-1} & \multicolumn{2}{c}{GSO} & \multicolumn{2}{c}{Sintel} & \multicolumn{2}{c}{DDAD} & \multicolumn{2}{c}{DIODE} & \multicolumn{2}{c}{Spring} & \multicolumn{2}{c}{HAMMER} \\

          & {\textit{\small{AbsRel}}}$\downarrow$ & {\small{{$\delta_1$}}}$\uparrow$ & {\small{\textit{AbsRel}}}$\downarrow$ 
          & {\small{$\delta_1$}}$\uparrow$ & 
        {\small{\textit{AbsRel}}}$\downarrow$ & {\small{$\delta_1$}}$\uparrow$ 
          & {\small{\textit{AbsRel}}}$\downarrow$ & 
          {\small{$\delta_1$}}$\uparrow$ & 
        {\small{\textit{AbsRel}}}$\downarrow$ & 
          {\small{$\delta_1$}}$\uparrow$
          & {\textit{\small{AbsRel}}}$\downarrow$ & {\small{{$\delta_1$}}}$\uparrow$ & {\small{\textit{AbsRel}}}$\downarrow$ 
          & {\small{$\delta_1$}}$\uparrow$ & 
        {\small{\textit{AbsRel}}}$\downarrow$ & {\small{$\delta_1$}}$\uparrow$ 
          & {\small{\textit{AbsRel}}}$\downarrow$ & 
          {\small{$\delta_1$}}$\uparrow$ & 
        {\small{\textit{AbsRel}}}$\downarrow$ & 
          {\small{$\delta_1$}}$\uparrow$ \\
    
        \hline
        {Marigold~\cite{ke2024repurposing}}        
        & 5.20 & 97.32 & 10.17 & 90.24 & 14.53 & 82.11 & 5.10 & 96.80 & 2.64 & 99.88 & 46.20 & 63.76 & 17.75 & 77.02 & 9.64 & 90.49 & 92.13 & 46.22 & 9.20 & 92.50 \\
        
        {MoGe~\cite{wang2025moge}}
        & \textbf{3.41} & {98.49} & 4.99 & {98.20} & \textbf{3.44} & {98.75} & {3.39} & 98.11 & \underline{0.96} & \textbf{99.99} & \underline{34.22} & 72.32 & \underline{10.08} & \underline{91.84} & \underline{5.00} & 95.80 & \textcolor{gray!50}{16.65} & \textcolor{gray!50}{86.35} & {3.31} & 98.46 \\
        {MoGe-2~\cite{wang2025moge2accuratemonoculargeometry}}
        & \underline{3.42} & 98.47 & \underline{4.80} & 98.19 & \underline{3.49} & \underline{98.78} & \textbf{2.69} & \textbf{98.90} & {0.97} & \textbf{99.99} & 38.40 & {72.56} & 10.39 & 91.28 & \textbf{4.72} & \textbf{96.39} & \underline{55.27} & \underline{66.90} & \underline{3.09} & {99.51} \\
        
        {FoundationGeo~\cite{liu2026foundationgeo}}
        & {3.56} & \textbf{98.58} & {4.85} & \underline{98.25} & {3.50} & \textbf{98.86} & \underline{2.93} & \underline{98.77} & {1.37} & \textbf{99.99} & \textbf{28.55} & \textbf{74.10} & \textbf{9.37} & \textbf{92.54} & {5.43} & {95.71} & \textcolor{gray!50}{16.66} & \textcolor{gray!50}{86.52} & 
        \textbf{2.53} & \textbf{99.66} \\

        {DepthAnything~\cite{yang2024depth}}
        & 7.32 & 95.57 & 13.37 & 81.59 & 7.57 & 94.35 & 6.62 & 96.26 & 2.05 & 99.98 & 39.97 & 71.08 & 16.06 & 81.01 & 8.78 & 93.20 & {69.53} & 58.14 & 7.80 & 96.75 \\
        {DepthAnythingV2~\cite{depth_anything_v2}}
        & 7.07 & 95.77 & 12.49 & 83.65 & 7.79 & 94.28 & 6.47 & 96.87 & 1.97 & 99.98 & 38.93 & 69.70 & 15.95 & 81.49 & 8.32 & 92.83 & 78.59 & 58.18 & 8.43 & 96.55 \\

        {DepthAnythingV3~\cite{depthanything3}}
        & 3.88 & 98.32 & 7.24 & 95.56 & 5.97 & 96.16 & {3.22} & 98.63 & 1.03 & \textbf{99.99} & 37.65 & {73.00} & 16.81 & 78.96 & 5.91 & 95.05 & \textbf{50.25} & \textbf{69.80} & 3.50 & {99.54} \\

        {VGGT~\cite{wang2025vggt}}
        & 3.52 & 98.28 & 9.49 & 90.87 & 4.52 & 96.68 & 4.62 & 96.48 & \textbf{0.84} & \textbf{99.99} & 48.93 & 66.15 & 17.71 & 77.75 & 7.92 & 91.98 & 88.34 & 59.28 & 3.94 & 97.95 \\
        {UniDepth~\cite{piccinelli2024unidepth}}
        & {3.79} & \underline{98.51} & \textbf{4.03} & \textbf{98.81} & 5.56 & 97.14 & 4.14 & 98.11 & 2.56 & 99.92 & 58.78 & 59.10 & {10.24} & {91.33} & 6.69 & 94.84 & 80.45 & 55.41 & 3.69 & 99.08 \\
        {UniDepthV2~\cite{piccinelli2025unidepthv2universalmonocularmetric}}
        &  6.08 & 96.50 & 8.46 & 93.37 & 9.48 & 91.27 & 6.81 & 94.53 & 2.50 & 99.79 & 58.50 & 54.20 & 17.69 & 78.03 & 12.02 & 86.74 & 112.20 & 41.51 & 12.27 & 82.54 \\
        {DepthPro~\cite{bochkovskii2024depth}}
        & 4.18 & 98.15 & 6.76 & 96.23 & 7.27 & 94.90 & 3.74 & 98.19 & 1.49 & \textbf{99.99} & 51.18 & 71.30 & 21.57 & 74.04 & 7.41 & 93.73 & 80.01 & 54.68 & 3.48 & \underline{99.65} \\
        {ZeroDepth~\cite{guizilini2023zeroshotscaleawaremonoculardepth}}
        & 6.03 & 96.39 & 7.61 & 94.74 & 13.24 & 85.07 & 6.95 & 93.91 & 4.95 & 99.41 & 66.04 & 51.10 & 17.06 & 78.69 & 19.09 & 79.98 & 152.30 & 34.38 & 14.74 & 78.82 \\
        {Metric3D~\cite{yin2023metric3d}}
        & 7.81 & 93.85 & 6.89 & 95.72 & 11.85 & 86.07 & 7.20 & 94.66 & 5.05 & 99.36 & 77.00 & 48.75 & \textcolor{gray!50}{18.86} & \textcolor{gray!50}{74.47} & 11.25 & 89.08 & 111.35 & 42.48 & 12.32 & 85.84 \\
        {Metric3DV2~\cite{Hu_2024}}
        & 6.77 & 94.07 & 5.43 & 98.07 & 5.51 & 96.49 & 4.34 & {98.26} & 1.78 & 99.98 & 38.59 & \underline{73.56} & \textcolor{gray!50}{10.87} & \textcolor{gray!50}{90.27} & 5.48 & \underline{96.32} & 84.61 & 55.59 & 3.83 & 98.57 \\
        {GeoWizard~\cite{fu2024geowizard}}
        & 5.12 & 97.41 & 9.68 & 91.41 & 9.02 & 92.06 & 4.91 & 97.11 & 1.75 & 99.97 & 45.22 & 67.72 & 21.13 & 71.59 & 8.85 & 91.86 & 93.39 & 54.17 & 4.38 & 98.47 \\
        {LeReS~\cite{yin2021learning}}
        & 6.75 & 95.54 & 12.61 & 83.61 & 11.69 & 87.53 & 7.30 & 94.69 & 4.14 & 99.68 & 58.49 & 56.40 & 19.43 & 73.01 & 13.82 & 84.95 & 104.48 & 48.08 & 8.44 & 93.88 \\
        {Lotus~\cite{he2024lotus}}
        & 8.26 & 94.30 & 8.46 & 93.56 & 10.24 & 91.98 & 8.56 & 93.87 & 3.04 & 99.95 & 46.25 & 66.85 & 15.51 & 81.39 & 10.54 & 90.26 & 70.59 & 57.90 & 8.74 & 95.13 \\
        {MiDas3.1~\cite{birkl2023midas}}
        & 7.32 & 95.53 & 12.23 & 84.52 & 9.39 & 91.77 & 6.98 & 95.72 & 2.28 & 99.97 & {36.76} & 70.61 & 18.19 & 77.25 & 10.14 & 90.74 & 82.02 & {59.15} & 8.78 & 94.58 \\
        {DPT~\cite{ranftl2021vision}}
        & 9.91 & 91.30 & 14.58 & 79.84 & 11.35 & 89.61 & 8.13 & 94.14 & 3.04 & 99.92 & 42.15 & 63.24 & 21.99 & 69.26 & 11.48 & 88.39 & 75.08 & 52.14 & 10.02 & 91.60 \\
        {PPD~\cite{xupixel}}
        & 5.61 & 96.51 & 22.22 & 66.58 & 12.70 & 85.79 & 6.75 & 94.42 & 1.80 & 99.98 & 47.88 & 64.64 & 26.73 & 65.33 & 11.83 & 86.65 & 99.28 & 52.87 & 3.99 & 99.35 \\
        {Diffusion-E2E-FT~\cite{garcia2025fine}}
        & 4.91 & 97.43 & 7.78 & 95.06 & 6.49 & 95.13 & 4.44 & 97.58 & 2.24 & 99.95 & 37.62 & 69.45 & 14.63 & 83.91 & 8.59 & 92.34 & 89.68 & 53.73 & 4.04 & 98.46 \\
        {GenPercept~\cite{xu2024matters}}
        & 4.93 & 97.48 & 8.60 & 93.72 & 6.84 & 95.04 & 4.84 & 97.48 & 1.97 & 99.98 & 40.50 & 69.58 & 16.14 & 80.76 & 9.28 & 90.47 & 96.34 & 53.81 & 3.50 & 99.36 \\

    \end{tabular}}
\noindent\rule{\linewidth}{0.5pt} %
  \adjustbox{width={\linewidth},keepaspectratio}{
    \begin{tabular}{l|cc|cc|cc|cc|cc|cc|cc|cc|cc|cc}

        \multirow{2}{*}{\textbf{Disparity Align}} & \multicolumn{2}{c}{NYUv2} & \multicolumn{2}{c}{KITTI} & \multicolumn{2}{c}{ETH3D} & \multicolumn{2}{c}{iBims-1} & \multicolumn{2}{c}{GSO} & \multicolumn{2}{c}{Sintel} & \multicolumn{2}{c}{DDAD} & \multicolumn{2}{c}{DIODE} & \multicolumn{2}{c}{Spring} & \multicolumn{2}{c}{HAMMER} \\
          & {\textit{\small{AbsRel}}}$\downarrow$ & {\small{{$\delta_1$}}}$\uparrow$ & {\small{\textit{AbsRel}}}$\downarrow$ 
          & {\small{$\delta_1$}}$\uparrow$ & 
        {\small{\textit{AbsRel}}}$\downarrow$ & {\small{$\delta_1$}}$\uparrow$ 
          & {\small{\textit{AbsRel}}}$\downarrow$ & 
          {\small{$\delta_1$}}$\uparrow$ & 
        {\small{\textit{AbsRel}}}$\downarrow$ & 
          {\small{$\delta_1$}}$\uparrow$
          & {\textit{\small{AbsRel}}}$\downarrow$ & {\small{{$\delta_1$}}}$\uparrow$ & {\small{\textit{AbsRel}}}$\downarrow$ 
          & {\small{$\delta_1$}}$\uparrow$ & 
        {\small{\textit{AbsRel}}}$\downarrow$ & {\small{$\delta_1$}}$\uparrow$ 
          & {\small{\textit{AbsRel}}}$\downarrow$ & 
          {\small{$\delta_1$}}$\uparrow$ & 
        {\small{\textit{AbsRel}}}$\downarrow$ & 
          {\small{$\delta_1$}}$\uparrow$ \\
        
        \hline
        {DepthAnything~\cite{yang2024depth}} 
        & {4.32} & {98.26} & {7.89} & {96.91} & {6.13} & {98.01} & {4.21} & {97.61} & {1.54} & {99.99} & 50.58 & {77.89} & 19.33 & {88.99} & {11.15} & {95.54} & {54.30} & {73.01} & 5.59 & 97.95 \\
        
        {DepthAnythingV2~\cite{depth_anything_v2}}
        & {4.37} & {98.08} & 9.54 & {96.27} & 19.98 &{97.66} & {3.70} & {98.22} & {1.18} &{99.99} & 55.41 & {74.29} & {17.10} & {87.81} & 13.91 & {95.98} & 66.97 & {68.92} & {5.00} & {99.05} \\
        {DPT~\cite{ranftl2021vision}}
        & 9.94 & 92.93 & 9.63 & 91.76 & 11.06 & 92.72 & 7.14 & 94.52 & 2.84 & 99.92 & {49.98} & 67.02 & 30.16 & 79.38 & 15.00 & 92.07 & {65.18} & 65.95 & 8.21 & 94.01 \\
        {Lotus~\cite{he2024lotus}}
        & 4.69 & 97.76 & {6.70} & 94.87 &{7.84} & 95.54 & 4.42 & 97.41 & 2.37 & 99.95 & {35.85} & 71.03 & {12.74} & 84.98 & {11.83} & 93.91 & 105.31 & 67.23 & {3.27} &{99.30} \\
        {MiDas3.1~\cite{birkl2023midas}}
        & 9.75 & 91.52 & 35.42 & 39.10 & 19.89 & 73.74 & 12.03 & 85.66 & 2.11 & 99.97 & 57.25 & 57.94 & 37.27 & 49.62 & 18.35 & 78.76 & 115.53 & 45.46 & 8.16 & 94.32 \\
    \bottomrule
    \end{tabular}}
\end{table*}

\subsection{Benchmark Performance}

Since different methods often adopt distinct evaluation protocols-- such as using different benchmark datasets or scale calibration strategies-- the results reported in individual papers are not directly comparable. To ensure a fair and consistent assessment, we standardize the evaluation under a unified framework.

First, we evaluate all methods on a common benchmark dataset. Specifically, we adopt the benchmark introduced by MoGe-2~\cite{wang2025moge2accuratemonoculargeometry}, which includes 10 datasets across multiple domains (e.g., indoor, street views, object scans and synthetic animations). This benchmark excludes ambiguous regions such as reflective surfaces and applies a consistent evaluation protocol to assess the predicted depth maps.  

Second, to align the predicted depths with ground-truth values, we apply a uniform affine-invariant alignment procedure, calculating a scale and shift using least-square fitting. We categorize the evaluated methods into three types:

\begin{enumerate}
\item Metric depth estimation:  UniDepth~\cite{piccinelli2024unidepth}, UniDepthV2~\cite{piccinelli2025unidepthv2universalmonocularmetric}, 
FoundationGeo~\cite{liu2026foundationgeo},
DepthPro~\cite{bochkovskii2024depth}, ZeroDepth~\cite{guizilini2023zeroshotscaleawaremonoculardepth}, Metric3D~\cite{yin2023metric3d},  Metric3DV2~\cite{Hu_2024} and MoGe-2~\cite{wang2025moge2accuratemonoculargeometry};
\item Relative depth estimation:  MoGe~\cite{wang2025moge},  Marigold~\cite{ke2024repurposing}, GeoWizard~\cite{fu2024geowizard}, LeReS~\cite{yin2021learning}, Diffusion-E2E-FT~\cite{garcia2025fine}, GenPercept~\cite{xu2024matters}; DepthAnythingV3~\cite{depthanything3},VGGT~\cite{wang2025vggt},
\item Disparity estimation: MiDaS v3.1~\cite{birkl2023midas} and DepthAnything~\cite{yang2024depth}, DepthAnythingV2~\cite{depth_anything_v2}, DPT~\cite{ranftl2021vision}, Lotus~\cite{he2024lotus}.
\end{enumerate}

Although metric depth models are designed to predict absolute depth values, they often suffer from scale inconsistency when directly compared to ground truth. Therefore, we apply affine alignment (scale + shift) even to metric methods prior to computing standard depth metrics. For relative depth models, which inherently lack absolute scale, we apply the same affine-invariant alignment. Disparity-based models are similarly aligned both in depth or disparity space. 
By standardizing all methods using affine-invariant alignment, we ensure a fair comparison that focuses solely on depth distribution accuracy, independent of scale and calibration variations. 

To ensure optimal performance, we follow each method’s official data preprocessing pipeline and evaluate their best-performing open-source implementations. We report the following commonly used metrics:

\begin{enumerate}
    \item Absolute Mean Relative Error($\text{AbsRel}$). Defined as $\frac{1}{M}|a_i - y^*_i| / y^*_i$,
    \item $\delta_1$ Accuracy. Defined as the proportion of pixels satisfying $\max(a_i/y^*_i, y^*_i/a_i) < 1.25$, 
\end{enumerate}
where $a_i$ is the aligned predicted depth, $y^*_i$ is the ground-truth depth, and $M$ is the total number of valid pixels.

\begin{figure*}[!htb]
    \centering
    \includegraphics[width=1.0\linewidth]{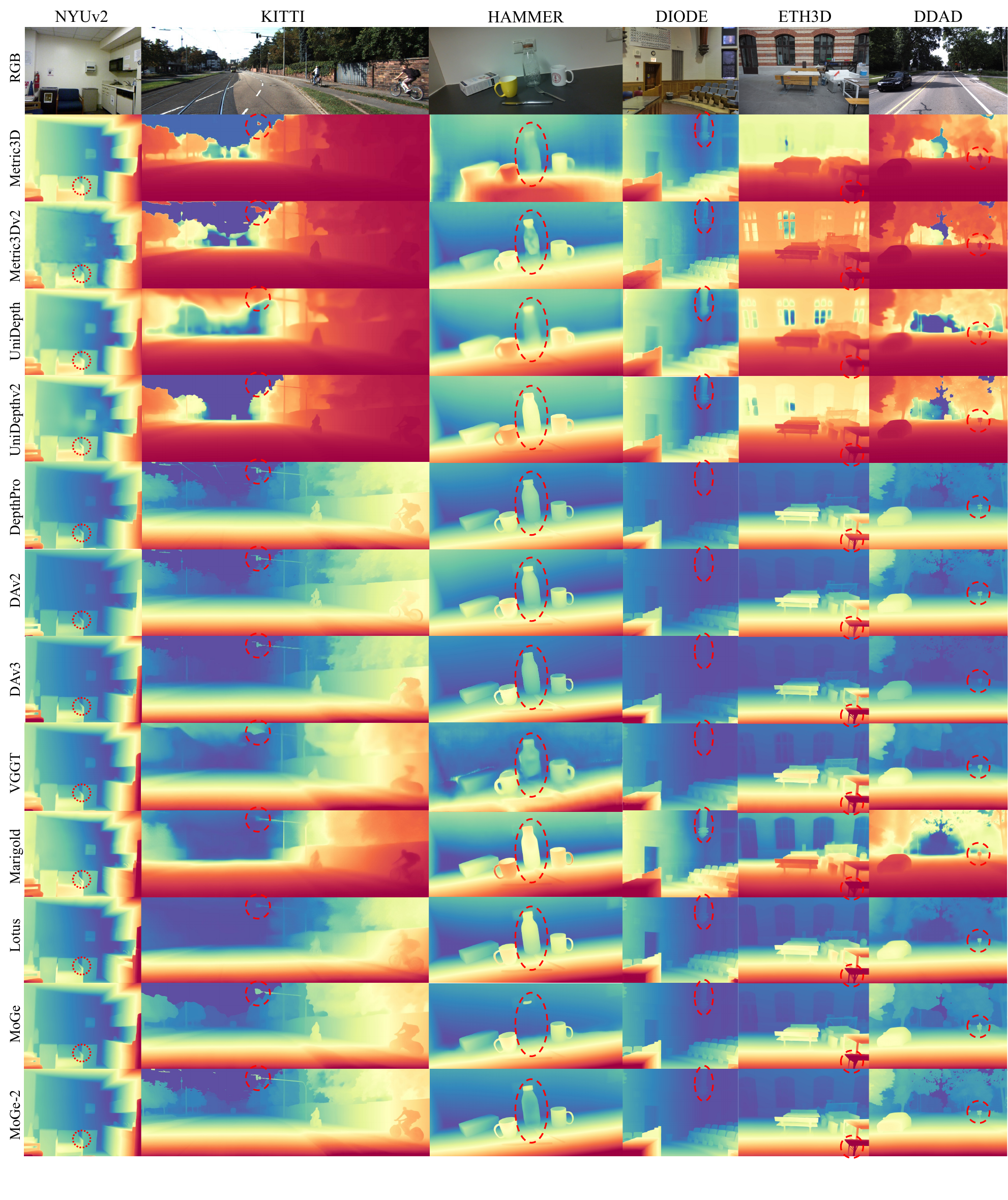}
    \caption{Qualitative visualization of depth estimation results from representative models~\cite{yin2023metric3d, Hu_2024, piccinelli2024unidepth, piccinelli2025unidepthv2universalmonocularmetric, bochkovskii2024depth, depth_anything_v2, ke2024repurposing, he2024lotus, wang2025moge, wang2025moge2accuratemonoculargeometry,wang2025vggt,depthanything3} across diverse scenarios, including indoor~\cite{Silberman2012IndoorSA, diode_dataset}, outdoor~\cite{8954208}, driving~\cite{geiger2013vision, packnet}, and object-level scenes~\cite{jung2023importanceaccurategeometrydata}. Despite some methods showed strong quantitative results, discrepancies remain between numerical performance and visual quality, particularly regarding boundary sharpness and detail preservation.}
    \label{fig:visualization}
\end{figure*}

\subsection{Extensions to Videos}
Video depth estimation introduces unique challenges beyond single-image prediction: 
\begin{itemize}
\item Temporal consistency is critical-- naïve frame-by-frame models often yield flickering results due to the lack of cross-frame awareness, leading to inconsistent scale across frames; 
\item Data scarcity is another bottleneck, as high-quality, temporally dense depth annotations for videos are difficult to obtain; 
\item Estimating depth over long sequences requires models that are both computationally efficient and temporally stable. 
\end{itemize}

\begin{figure}[t]
    \centering
    \includegraphics[width=1.0\linewidth]{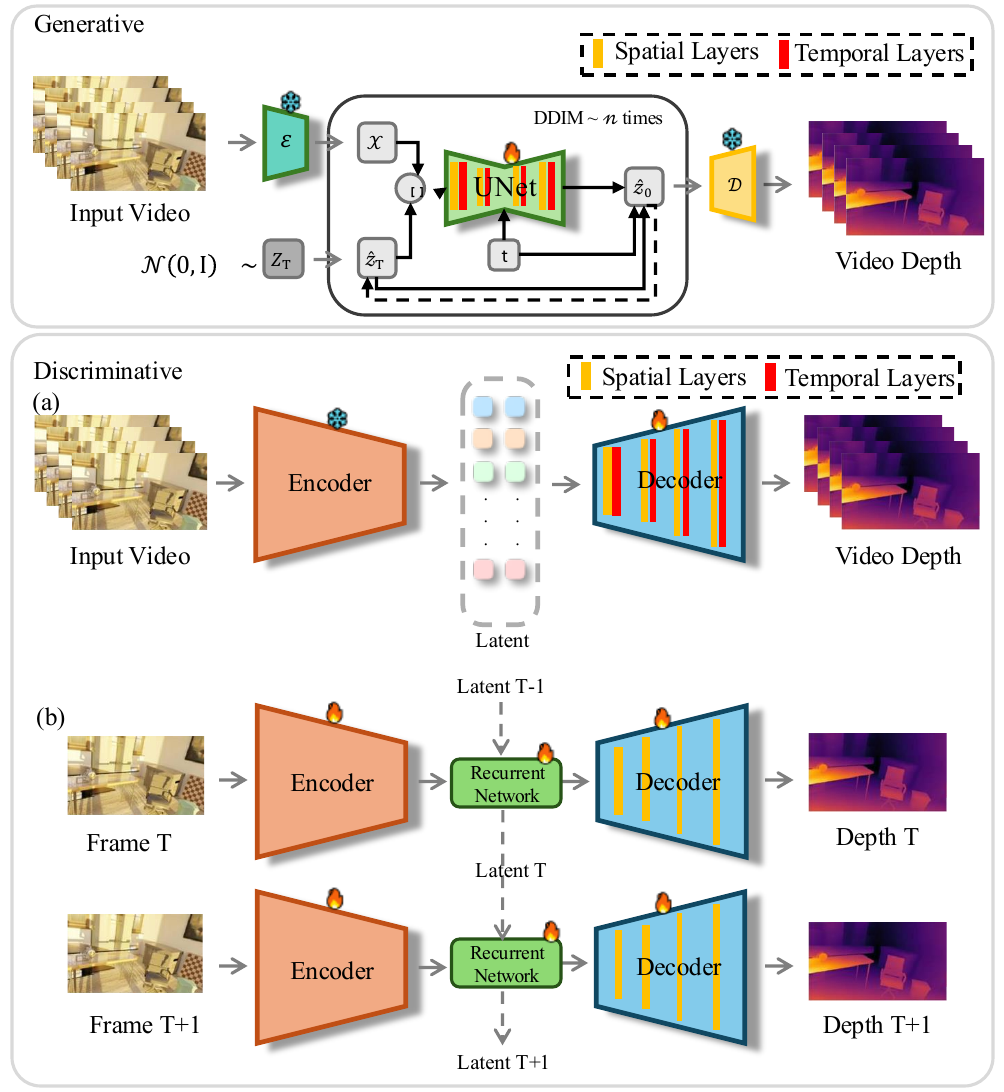}
    \caption{Structure of video depth estimation. Unlike image depth estimation models, which use spatial layers to process each image, video depth estimation models additionally introduce temporal layers or utilize previous latent states via a recurrent neural network to ensure consistency across frames.}
    \label{fig:Extensions_to_video}
\end{figure}

As shown in Fig.~\ref{fig:Extensions_to_video}, recent methods~\cite{hu2025-DepthCrafter, yang2024depthanyvideo, video_depth_anything, shao2024learningtemporallyconsistentvideo, sun2025unigeo, chou2025flashdepth,lyu2026streamingdepth} can be broadly classified into two categories: (1) generative, diffusion-based approaches and (2) discriminative models equipped with temporal consistency modules, encompassing both offline and streaming-based designs. 

\paragraph{Generative Video Diffusion Model} {Generative diffusion-based models}-- including DepthCrafter \cite{hu2025-DepthCrafter}, ChronoDepth \cite{shao2024learningtemporallyconsistentvideo}, and Depth Any Video \cite{yang2024depthanyvideo}-- leverage pretrained video diffusion priors to jointly model spatial detail and temporal coherence. DepthCrafter \cite{hu2025-DepthCrafter} processes up to 110 frames in parallel and uses overlap-stitching for long videos. ChronoDepth \cite{shao2024learningtemporallyconsistentvideo} introduces a two-stage training regime, learning spatial representations before injecting temporal consistency via cross-frame attention. Depth Any Video \cite{yang2024depthanyvideo} scales training with 40K+ synthetic clips and employs interpolation to infer depth over up to 150 frames. These models operate without pose or flow supervision, generalize effectively in zero-shot settings, and produce high-fidelity depth maps with strong temporal stability. However, they typically do not support streaming inference, and their memory usage grows with the number of input frames. 

\paragraph{Discriminative (Streamable) Models} Discriminative models such as Video Depth Anything \cite{video_depth_anything} (Fig.~\ref{fig:Extensions_to_video} (a)) and FlashDepth \cite{chou2025flashdepth} (Fig.~\ref{fig:Extensions_to_video} (b)), aim to provide video depth estimation without diffusion-based generation. Both build upon Depth Anything V2 as the core monocular backbone but introduce temporal modeling to stabilize predictions over time. Video Depth Anything \cite{video_depth_anything} integrates a lightweight spatio-temporal head and a temporal gradient consistency loss to enforce temporal smoothness. However, it still faces challenges in handling streaming videos. 
Further, FlashDepth \cite{chou2025flashdepth} takes a more architectural approach to temporal modeling. It introduces a dual-stream design, combining a fast high-resolution stream and a robust low-resolution stream, fused via cross-attention. A recurrent module (Mamba) aligns temporal features across frames. This architecture enables real-time depth estimation on full 2K resolution videos (2044×1148), offering sharper boundaries and strong temporal stability without batching or offline processing and naturally supporting for streaming processing. 
More recently, DyFN~\cite{lyu2026streamingdepth} attributes streaming inconsistency to scale--shift drift caused by fluctuations in latent feature statistics, rather than inaccurate per-frame geometry. It introduces a lightweight causal recurrent normalization module to stabilize these statistics over time. By training only DyFN while freezing the backbone, the method preserves single-frame accuracy and improves temporal stability by up to 14\%.

\subsection{Extensions to Any Camera}
Most depth foundation models are developed and evaluated under rectified perspective assumptions, whereas real-world systems frequently use {arbitrary projection models} and {large fields of view} (FoV), e.g., fisheye and $360^\circ$ panoramas. Naively transferring perspective-centric pipelines often fails due to unmodeled spherical distortions and the loss of global context. Moving from pinhole settings to {any-camera} depth (and 3D) estimation introduces several additional challenges:
\begin{itemize}
    \item \textbf{Geometry mismatch:} the pixel-to-ray mapping changes across camera models, making learned pinhole priors brittle under wide-FoV distortions.
    \item \textbf{Context--distortion trade-off:} cropping/undistortion either discards long-range context or introduces spatially varying warps that violate common inductive biases.
    \item \textbf{Data scarcity:} paired metric depth for fisheye/panoramic imagery is far less available than perspective RGB-D/LiDAR supervision.
    \item \textbf{FoV/resolution variability:} canonical spherical/ERP representations lead to uneven sampling densities and resolution mismatches, complicating stable training and deployment.
\end{itemize}

Recent progress~\cite{pintore2024deep,piccinelli2025unik3d,guo2025depth,cao2025panda,li2025da2,lin2025depth} follows a shared recipe: combine {geometry-aligned canonicalization or parameterization} with {data scaling/adaptation}, instead of relying on entirely new backbones. UniK3D~\cite{piccinelli2025unik3d} makes camera variability explicit by predicting a model-independent {pencil-of-rays} in spherical form and a per-ray {radial distance}, enabling metric 3D/depth across wide-FoV cameras. Depth Any Camera (DAC)~\cite{guo2025depth} keeps training strictly on perspective RGB-D yet improves generalization by mapping inputs into a canonical ERP space with pitch-aware conversion, FoV alignment, and multi-resolution augmentation. PanDA~\cite{cao2025panda} adapts Depth Anything to panoramas via teacher--student self-training on large-scale {unlabeled} panoramas, regularized by Möbius transformation-based spatial augmentation to enforce spherical consistency. Depth Any Panoramas (DAP)~\cite{lin2025depth} pushes the foundation-model regime for panoramas through a data-in-the-loop engine (public + synthetic + web panoramas) with multi-stage pseudo-label curation, complemented by panoramic-specific heads/objectives (e.g., range masking and geometry/sharpness-oriented optimization) for stable metric prediction across diverse distances. Finally, DA$^{2}$~\cite{li2025da2} couples a scalable panoramic data curation engine (perspective-to-panorama generation) with a sphere-aware ViT that encodes spherical coordinates, yielding an end-to-end panoramic estimator with strong zero-shot generalization and improved efficiency over perspective-splitting pipelines. Overall, these methods suggest that ``any-camera'' depth hinges on {explicit geometric alignment} and {scalable supervision/regularization} that lets existing foundation priors extend beyond the pinhole regime.

\subsection{Summary}
In summary, recent depth foundation models demonstrate that scaling up heterogeneous LiDAR, RGB-D, synthetic, and pseudo-labeled data, together with strong pre-trained encoders, is key to achieving zero-shot robustness and detail-preserving depth. Discriminative ViT-based architectures~\cite{wang2025moge2accuratemonoculargeometry,wang2025moge,bochkovskii2024depth,depth_anything_v2,yang2024depth,piccinelli2024unidepth,piccinelli2025unidepthv2universalmonocularmetric,ren2026anydepth}, spanning relative and metric predictors as well as point-based representations, provide efficient and accurate estimation when combined with affine-invariant objectives, camera-aware designs, and lightweight global scale heads. Generative, diffusion-based approaches~\cite{ke2024repurposing, fu2024geowizard, xu2024matters, garcia2025fine, zhang2024betterdepth, he2024lotus, song2026depthmaster} complement them by injecting powerful image priors and excelling at fine-grained geometry, and are increasingly distilled into one-step or hybrid refiners that close the gap in efficiency. Extending these ingredients to videos via video diffusion models~\cite{hu2025-DepthCrafter,yang2024depthanyvideo,shao2024learningtemporallyconsistentvideo} and lightweight temporal modules~\cite{video_depth_anything,chou2025flashdepth} further enables temporally consistent, streamable depth over long and high-resolution sequences, pointing toward unified depth foundation models across relative/metric and image/video settings. Finally, any-camera extensions~\cite{piccinelli2025unik3d,guo2025depth,cao2025panda,lin2025depth,li20252} suggest that robust non-pinhole generalization is driven by {explicit geometric alignment} (e.g., spherical ray--radial or ERP canonicalization) plus {scalable panoramic supervision/regularization}, enabling strong zero-shot transfer to fisheye and $360^\circ$ cameras.

\section{Downstream Applications of Depth Estimation} 
\label{sec:downstream}
\begin{figure*}[htbp]
    \centering
    \includegraphics[width=0.95\linewidth]{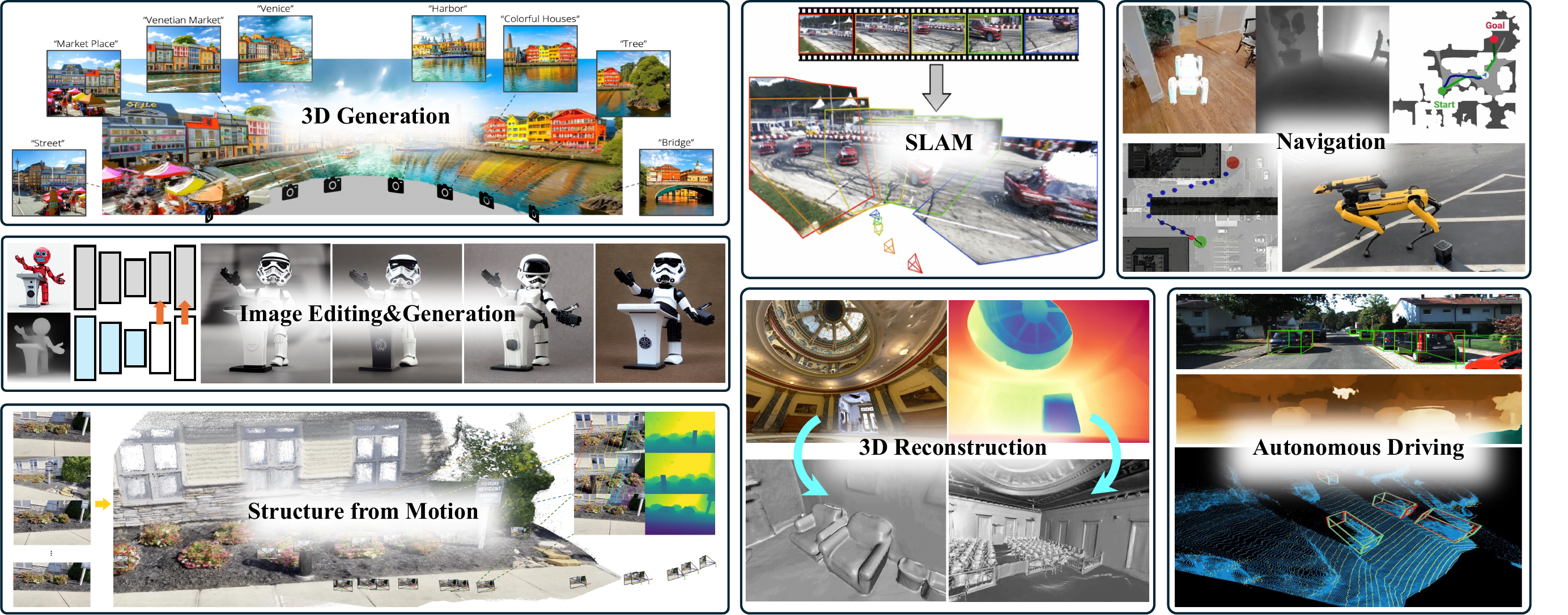}
    \caption{Depth estimation, a fundamental challenge in computer vision, plays a vital role in enabling numerous critical downstream tasks, such as 3D generation~\cite{yu2025wonderworld}, image editing\&generation~\cite{zhang2023adding}, structure from motion~\cite{wang2025vggt}, simultaneous localization and mapping (SLAM)~\cite{li2025megasam}, 3D reconstruction~\cite{yu2022monosdf}, autonomous driving~\cite{wang2019pseudo}, and robotic navigation~\cite{truong2024indoorsim}. }
    
    \label{fig:applications}
\end{figure*}

High-quality depth maps from monocular images unlock a wide array of downstream applications in computer vision and graphics. 
Fig.~\ref{fig:applications} illustrates several important and commonly encountered applications closely associated with depth estimation.
In this section, we outline several categories of applications that have benefited from advances in depth estimation, particularly emphasizing how the recent models reviewed above contribute to each. 

\subsection{SLAM and 3D/4D Scene Reconstruction}

Depth information is fundamental to 3D reconstruction from images. Classical methods such as Structure-from-Motion (SfM)\cite{schonberger2016structure}, Multi-View Stereo (MVS)\cite{schoenberger2016mvs}, and SLAM~\cite{durrant2006simultaneous} reconstruct scenes by triangulating image correspondences, requiring sufficient camera motion and precise initialization-- both of which benefit from accurate depth cues.

Prior to the rise of deep learning, depth sensors played a central role in fusion-based SLAM systems. KinectFusion~\cite{izadi2011kinectfusion} and its extensions~\cite{chen2013scalable,kahler2015very,niessner2014combining} introduced volumetric fusion for dense mapping. Subsequent systems like RGB-D SLAM~\cite{endres20133}, ElasticFusion~\cite{whelan2016elasticfusion}, and BundleFusion~\cite{dai2017bundlefusion} leveraged active depth sensing for real-time, high-quality 3D reconstruction.

\paragraph{SLAM} As depth sensors face limitations in resolution and robustness, deep learning has emerged to refine or replace sensor-based depth. Early works improved depth maps via co-training with discriminators~\cite{wu2019spatial} and multi-view fusion~\cite{bae2022multi}. Other approaches~\cite{yin2017scale,yang2018deep} recovered metric depth from predicted relative depths. CNN-SLAM~\cite{tateno2017cnn,loo2019cnn} was among the first to integrate CNN-based monocular depth estimators into SLAM, replacing depth sensors. More recently, MegaSaM~\cite{li2025megasam} scaled flow-based SLAM~\cite{teed2021droid} by leveraging DepthAnything-V2~\cite{depth_anything_v2} and calibrating it with UniDepth~\cite{piccinelli2024unidepth}, enhancing SLAM accuracy on large-scale datasets. 
Metric monocular depth estimation has also enabled unified SLAM and mapping in NeRF and 3DGS-based systems. IMAP~\cite{sucar2021imap} first demonstrated NeRF as a SLAM map representation, where depth plays a key role in initialization and optimization. NICE-SLAM~\cite{zhu2022nice} introduced coarse-to-fine rendering loss for robust bundle adjustment, and NICER-SLAM~\cite{zhu2024nicer} eliminated depth sensors altogether, relying on monocular cues. In parallel, 3DGS-based SLAM systems~\cite{matsuki2024gaussian,yugay2023gaussian,yan2024gs} adopted monocular depth for regularization and isotropic Gaussian constraints.

\paragraph{3D/4D Scene Reconstruction and Rendering} Beyond SLAM, monocular depth estimation has become key for geometry reconstruction~\cite{gao2025more,fang2026dens3r} and scene generation. VGGT~\cite{wang2025vggt} extends DUSt3R~\cite{wang2024dust3r} with a transformer-based feedforward SfM model, using multi-task training and a depth estimation head to jointly enhance SfM and depth learning. DAv3~\cite{depthanything3} uses a lightweight vanilla DINO transformer backbone with layer-level feature transfer and a unified depth-ray prediction target, eliminating the need for complex multi-task objectives while yielding spatially consistent geometry. MapAnything~\cite{keetha2025mapanything} further demonstrates that by conditioning a unified transformer on auxiliary geometric priors such as poses and camera intrinsics, we can reinforce monocular depth cues and directly regress a globally consistent, metric-scaled 3D scene. Methods like Prometheus~\cite{yang2025prometheus} and SplatFlow~\cite{go2025splatflow} encode multi-view inputs and estimated monocular depths into latent spaces, where flow~\cite{liuflow} or diffusion models~\cite{ho2020denoising} are trained for 3D scene generation~\cite{kerbl3Dgaussians}. 

Neural rendering approaches, particularly NeRF~\cite{mildenhall2021nerf}, have shown that integrating monocular depth cues improves scene reconstruction accuracy and robustness. Models like MonoSDF~\cite{yu2022monosdf} and Neuralangelo~\cite{li2023neuralangelo} incorporate depth priors to constrain geometry. 
To effectively leverage depth order information and address the metric ambiguity inherent in monocular depth models, SparseNeRF~\cite{wang2023sparsenerf} introduces a ranking loss to regularize the NeRF geometry.
In 3D Gaussian Splatting (3DGS)~\cite{kerbl3Dgaussians}, now popular for its speed and rendering quality, depth supervision becomes even more critical, especially under sparse-view conditions~\cite{xu2025depthsplat,chung2024depth,sun2024splatter,wang2024shape,liu2024modgs,shi2025unisplat}, to guide the alignment and regularization of Gaussian primitives. In essence, these methods reveal a growing convergence between monocular depth estimation and NeRF-like generative representations: depth is increasingly used as an explicit supervisory signal or geometric prior to regularize neural scene models, especially under sparse or unconstrained observations.
This trend helps bridge the traditional “estimation” view and modern 3D generative modeling.  

Monocular depth estimation has matured into a powerful, geometry-aware signal that underlies both classical and learning-based SLAM, SfM, and neural reconstruction pipelines, enabling scalable, accurate, and sensor-free 3D mapping. It has also become a key cue for 4D reconstruction from in-the-wild monocular videos, supporting dynamic scene understanding with minimal sensing requirements.

\subsection{Image Editing and Content Creation}

Depth estimation has become a foundational tool in image editing and creative applications, enabling convincing 3D effects and spatially-aware manipulation from 2D inputs.

\paragraph{View Synthesis and Parallax Effects}
A single image with an estimated depth map can be re-rendered from new viewpoints to produce 3D photo effects by warping and inpainting disoccluded regions~\cite{shih20203d, jampani2021slide}. Popularized by platforms like Facebook, these effects rely on layered depth images or shallow meshes derived from monocular depth networks like MiDaS. For wider view changes, recent methods~\cite{fridman2023scenescape, hollein2023text2room, yu2025wonderworld} use progressive depth-guided warping and large-scale inpainting to reconstruct room-scale 3D scenes. Stereoscopic image synthesis~\cite{wang2024stereodiffusion, kopf2020one} uplifts monocular input into VR-ready left/right-eye pairs, while video extensions such as SVG~\cite{dai2024svg} and StereoCrafter~\cite{zhao2024stereocrafter} achieve temporally consistent depth-aware 3D video synthesis. These applications heavily depend on high-quality depth with sharp boundaries and accurate ordinal relations to minimize distortions and disocclusion artifacts.

\paragraph{Control for Generative Models}
Depth maps are now widely used to guide image generation. Models like ControlNet~\cite{zhang2023adding} enable depth-to-image pipelines that preserve scene geometry while allowing stylized or semantic modifications via prompts. For instance, a real-world image can be depth-estimated and transformed into a stylized winter scene while maintaining structural fidelity. Depth is also used to guide 3D-aware image/video synthesis~\cite{gu2025diffusion, wang2025cinemaster,li2024depthgan}, with generation quality strongly tied to depth accuracy. 
Foundation models significantly improve control and realism in such applications.

\paragraph{Computational Photography}
Depth supports advanced photography tasks, such as portrait mode (bokeh), where background blur is applied using depth to simulate shallow depth-of-field~\cite{wadhwa2018synthetic, peng2022bokehme}. Accurate depth around edges is critical to avoid visual artifacts. For relighting, depth enables geometric normal estimation~\cite{badino2011fast, izadi2011kinectfusion} and can be combined with material and lighting models for photorealistic rendering~\cite{pharr2023physically}. Models like GeoWizard~\cite{fu2024geowizard} jointly predict depth and normals, supporting neural relighting frameworks~\cite{el2021ntire} for effects like shadows and specular highlights.

\subsection{Robotic Navigation and Manipulation}
Monocular depth estimation has become a cornerstone of modern robotic systems, providing a lightweight, low-cost alternative to LiDAR and stereo cameras for 3D scene understanding. Depth foundation models transform RGB inputs into metric or relative depth maps, enabling robots to perceive obstacles, segment free space, and plan safe trajectories~\cite{cao2016exploiting, schon2023impact, wang2021domain}.

\paragraph{Navigation}

\par Monocular depth estimation has become a lightweight, cost-effective alternative to LiDAR and stereo rigs for map-less navigation~\cite{pei2021improved,zhang2025survey,wang2025mgnav}. Early efforts like ORB-SLAM~\cite{mur2015orb} reconstructed sparse 3D maps from a single moving camera, while Mancini \textit{et al.} \cite{mancini2016fast} demonstrated real-time depth prediction from RGB for obstacle avoidance. Recent works focus on fully onboard deployment: MonoNav\cite{simon2023mononav} enables micro-UAVs to estimate depth at 30 fps, Dang \textit{et al.} \cite{dang2023obstacle} introduced an FCN-based obstacle-avoidance strategy, and Lee \textit{et al.} \cite{lee2021deep} fused object detection and depth regression for UAVs in plantations. Integrated pipelines like Machkour \textit{et al.}~\cite{machkour2023monocular} combine MonoDepth with detection and segmentation for target-driven, obstacle-aware navigation.

\paragraph{Manipulation and Grasping}
Monocular depth estimation has also been applied to robotic manipulation and grasp planning. Recent studies~\cite{loss2021improving, atar2024optigrasp, guo2025monocular,qian2025geopredict} show that incorporating monocular depth into grasping pipelines improves real-time success rates and accurate 6-DoF pose estimation without additional sensors. While industrial systems typically use structured-light sensors for reliability, learned depth models can handle challenging conditions, such as glare or transparent objects, by leveraging shape priors~\cite{huang2025spatial, jain2023depth, ma2024sim}.

\paragraph{Autonomous Driving}
In autonomous driving, monocular depth estimation serves as a cost-effective alternative to LiDAR, with growing integration into perception pipelines. One strategy converts depth maps into pseudo-LiDAR point clouds to enable LiDAR-based 3D detection, significantly improving vehicle detection accuracy~\cite{wang2019pseudo, you2019pseudo}. Other approaches combine monocular depth modules with object detectors in multitask frameworks to jointly estimate object identity and distance~\cite{shen2020joint, qian2020end, wang2021plumenet,yan2026vg3s,shi2025drivex}. These models deliver dense, accurate depth in diverse environments, supporting next-generation autonomous systems.

\paragraph{Medical and Surgical Robotics}
In medical and surgical robotics, monocular depth estimation is essential for scene understanding when only endoscopic or microscopic views are available. Early methods addressed the lack of ground truth by training on synthetic data~\cite{mahmood2018unsupervised} or using self-supervised video learning to jointly estimate depth and camera pose without explicit labels~\cite{turan2018unsupervised}. More recently, medical depth foundation models~\cite{zeinoddin2024dares, lou2024surgical, li2024advancing, xu2024daua} built on the DepthAnything~\cite{yang2024depth} architecture have set new benchmarks in accuracy and robustness for robotic endoscopy.

\subsection{Beyond RGB: Multimodal and Other Uses}
Monocular depth estimation is increasingly used to complement other depth sensing modalities and to extend depth perception into specialized domains. 

\paragraph{Combine with Other Depth Sensing Techniques}
Monocular depth estimation enhances other sensing techniques by providing dense, semantic, and structurally coherent priors that address the limitations of traditional sensors. In FoundationStereo~\cite{wen2025foundationstereo}, monocular priors guide stereo matching to improve robustness in textureless or ambiguous regions. Prompt Depth Anything~\cite{wang2025depth} uses sparse LiDAR as a prompt, while dense monocular features help propagate and refine measurements into high-resolution depth maps. PriorDepth~\cite{lin2025prompting} leverages monocular predictions to complete and regularize noisy or sparse priors from LiDAR or SfM. These works show how monocular estimation offers semantic guidance that boosts the accuracy, coverage, and fidelity of hybrid depth systems.

\paragraph{Specialized Domains} 
Monocular depth estimation, initially developed for general scenes, is increasingly applied to specialized domains like underwater and medical imaging, where dense ground-truth depth is hard to obtain. Atlantis~\cite{zhang2024atlantis} uses diffusion-based synthetic data generation to train models on photorealistic underwater images with known geometry, enabling robust depth prediction without real annotations. In the medical domain ~\cite{han2024depthmedicalimagescomparative}, Depth Anything shows strong zero-shot generalization to endoscopic and laparoscopic scenes, providing a viable alternative to task-specific models when labeled data is scarce. These advances highlight the adaptability of monocular depth models through synthetic data, domain transfer, and prompt-based conditioning, expanding their utility in challenging sensing environments.

\section{Future Research Directions}
\label{sec:future}
Monocular depth estimation has made remarkable progress, yet several challenges and open research questions remain. We discuss some promising directions for future work, informed by the gaps observed in current literature and emerging needs in practical applications. 

\paragraph{Closing the Gap between Relative and Metric Depth}
While foundation models have narrowed the gap between relative and metric depth, achieving universal metric depth estimation remains an open challenge. Approaches like Metric3D~\cite{yin2023metric3d}, UniDepth~\cite{piccinelli2024unidepth}, and DepthPro~\cite{bochkovskii2024depth} demonstrate promising generalization, but further refinement is needed. One direction is improving camera adaptability-- developing models that can infer or adjust to intrinsic parameters beyond focal length (e.g., distortion, sensor size) would enhance cross-device reliability. Another promising idea is few-shot metric calibration: a model could produce relative depth by default, but globally adjust scale using a few metric reference points. Learning fast scale adaptation through meta-networks could enable easy per-device calibration with minimal user input.

\paragraph{High-Fidelity Depth and Uncertainty}
As depth models grow more accurate, there is rising interest in quantifying uncertainty to inform downstream tasks. Depth predictions are inherently ambiguous in textureless or distant regions, and per-pixel confidence or distributions over depth values would help assess reliability. Diffusion models can sample multiple plausible depth maps, but converting these into actionable statistics (e.g., mean and variance) remains an open challenge. Existing approaches like ensembles or Monte Carlo dropout~\cite{marsal2023monoprobselfsupervisedmonoculardepth} offer uncertainty estimates, but are impractical for large foundation models. An alternative is training models to output a small set of quantized hypotheses or parametric distributions (e.g., Gaussian) per pixel.

Another frontier is achieving ultra-high resolution and fine detail. While models like PatchFusion~\cite{li2024patchfusion} and DepthPro~\cite{bochkovskii2024depth} address high-res input, scaling to 4K or 8K remains limited by memory and edge preservation. Future directions include multi-scale transformers or implicit neural representations that produce continuous-resolution outputs. A persistent challenge is thin structure recovery-- even top models struggle with wires or branches. This may require targeted synthetic data (e.g., CAD-generated thin structures) or novel loss functions that enforce topological consistency.

\paragraph{Challenging Regions: Sky, Boundaries, and Non-Lambertian Surfaces}
Beyond average accuracy, persistent failure cases cluster in several hard regions.
\emph{Infinite-distance} areas (e.g., sky/horizon) provide weak geometric cues, so models may extrapolate arbitrary metric depth; treating these pixels as open-set (e.g., ``unknown/infinity'') and pairing predictions with calibrated uncertainty is a promising direction.
\emph{Depth discontinuities} at object boundaries are often over-smoothed, suggesting more boundary-aware training (edge/normal consistency, segmentation-guided refinement) and topology-preserving constraints for thin structures.
Finally, \emph{transparent/reflective} surfaces remain difficult due to non-Lambertian effects and because RGB-D/LiDAR supervision is frequently missing or corrupted~\cite{liu2026optigeo}; future work may combine robust learning under partial labels with targeted synthetic data and additional cues (e.g., polarization) to better handle these cases.

\paragraph{Temporal and Multi-View Consistency}
Most foundation models are trained on single images and lack built-in mechanisms for multi-view or temporal coherence, leading to flickering and inconsistency in video or multi-camera setups. A promising direction is to develop depth models that enforce temporal or multi-view consistency. This could involve training on videos with known depth using temporal losses, or introducing recurrent or transformer-based modules that propagate depth across frames. Alternatively, incorporating optical flow or feature tracking can help maintain depth stability on moving objects. To retain generalization from single-image training, a two-stage setup may be effective-- using a foundation model for initial depth, followed by a lightweight refinement module conditioned on temporal context. 

\paragraph{Domain Adaptation and Special Domains} 
Despite broad training data, specialized domains like medical imaging, satellite views, or underwater scenes often fall outside the scope of current models. Future work may explore domain adaptation to transfer foundation depth models with minimal supervision—e.g., adapting Depth Anything~\cite{yang2024depth} to underwater imagery using unsupervised methods guided by physics priors like light attenuation. Another promising direction is multi-modal depth estimation, incorporating inputs such as polarization or language. For instance, language prompts like ``the floor is 3 meters away'' can help resolve depth ambiguities~\cite{zeng2024wordepthvariationallanguageprior}, enabling models to reason about scale using object priors. Integrating such high-level knowledge—e.g., knowing typical car heights to calibrate predictions—opens the door to LLM-assisted depth models that blend perception with semantics.

\paragraph{Benchmarking and Evaluation Metrics}
As depth models advance, traditional metrics like RMSE may no longer reflect perceptual or structural quality. New evaluation protocols are needed to assess cross-dataset generalization and 3D consistency. Ongoing efforts promote diverse benchmarks, but expanding to continual evaluation streams-- where models are tested on a steady flow of unseen images or videos-- b   could better measure robustness. Additionally, assessing geometric plausibility, such as reprojecting predicted depth into multiple views for consistency, would encourage models to produce globally coherent 3D structure rather than optimizing only pointwise accuracy.

\paragraph{Integration with 3D Understanding}
Depth is one component of scene understanding, and integrating it with object recognition, segmentation, and 3D detection remains an open challenge. End-to-end systems that jointly predict depth and detect objects~\cite{zhang2025monodetrdepthguidedtransformermonocular} can mutually benefit—depth aids object scaling, while detection provides semantic grounding. Early multi-task learning efforts exist, but more work is needed to effectively fuse depth with other labels. Future foundation models may output holistic scene representations: layered depth maps with segmented objects, or even coarse 3D reconstructions—moving toward neural scene representations where monocular depth is one element of a unified 3D model.

We also anticipate growing synergy between depth and generative models. Generative AI can synthesize rare or challenging training samples (e.g., extreme weather) to fill data gaps. Conversely, depth predictions can guide 3D generative models like NeRFs~\cite{mildenhall2021nerf} or 3D GANs~\cite{goodfellow2014generativeadversarialnetworks}, for example, by initializing NeRFs to accelerate convergence. As these domains converge, the boundary between estimation and generation may blur, enabling systems that both perceive and imagine depth beyond visible scenes.

In conclusion, monocular depth estimation is moving toward ever more general, accurate, and integrated capabilities. The community is actively addressing current limitations, and we anticipate that future models will seamlessly provide dense, metric, and reliable depth for any image or video, becoming a cornerstone of 3D vision in everyday devices and complex AI systems alike.

\section{CONCLUSION}
\label{sec:conclusion}
Monocular depth estimation has rapidly evolved from initial learning-based methods to contemporary foundation models, achieving remarkable progress in accuracy, robustness, and generalization. Recent advancements have effectively bridged the gap between relative and metric depth estimation, integrating powerful vision transformer architectures, large-scale synthetic and pseudo-labeled datasets, and pretrained vision foundation models like DINOv2 and diffusion models. These developments have unlocked numerous practical applications across 3D reconstruction, visual SLAM, image editing, robotics, and multimedia content creation. Nevertheless, challenges remain, including enhancing temporal and multi-view consistency, addressing specialized domain adaptation, quantifying predictive uncertainty, and integrating comprehensive scene understanding. Future research in these directions promises to further solidify monocular depth estimation as a fundamental component of computer vision, enabling robust, scalable, and universally applicable depth perception capabilities.

\subsection*{\color{main}Availability of data and materials}
The evaluation datasets we used in the benchmark part are all publicly released datasets. Our code and scripts are available at \href{https://github.com/CVMI-Lab/Depth_Survey}{CVMI-Lab/Depth-Survey}.

\subsection*{\color{main}Competing interest}

The authors have no competing interests to declare that are relevant to the content of this article.

\subsection*{\color{main}Funding}

This work was supported in part by the Hong Kong Research Grant Council — Early Career Scheme (Grant No. 27209621), General Research Fund Scheme (Grant Nos. 17202422, 17212923, 17215025), and Theme-based Research Scheme (Grant No. T45-701/22-R), as well as by the Shenzhen Science and Technology Innovation Commission (Grant No. SGDX20220530111405040). Part of the research was conducted at the JC STEM Lab of Robotics for Soft Materials, funded by The Hong Kong Jockey Club Charities Trust.

\subsection*{\color{main}Acknowledgements}
The authors would like to thank DiDi AI Research. Part of the work presented in this survey was completed by Muxin and Xiaoyang during their time at the DiDi AI Research group.

\appendix




\bibliographystyle{CVMbib}
\bibliography{refs}


\begin{biography}[figures/authors/author1]{Muxin Liu} received his bachelor's degree from Southwest Jiaotong University in 2021 and master's degree from the University of Hong Kong in 2024. He is currently a Ph.D. student at the University of Hong Kong. His research interests include 3D computer vision and autonomous systems.
\end{biography}
\vspace*{1.2em}

\begin{biography}[figures/authors/author2]{Xiaoyang Lyu} received his bachelor's degree from Harbin Institute of Technology in 2019 and his master's degree from Zhejiang University in 2022. He is currently a PhD candidate at the University of Hong Kong. His research interests include neural rendering and reconstruction.
\end{biography}
\vspace*{1.2em}

\begin{biography}[figures/authors/author3]{Yang-Tian Sun} received his bachelor's degree in applied physics from the University of Science and Technology Beijing and master's degree in computer science from the Institute of Computing Technology, Chinese Academy of Sciences. He is currently a PhD candidate at the University of Hong Kong. His research interests include computer graphics and computer vision.
\end{biography}
\vspace*{1.2em}

\begin{biography}[figures/authors/author4]{Yi-Hua Huang} received his bachelor’s and master’s degrees from the University of Chinese Academy of Sciences. He is currently a Ph.D. student at the University of Hong Kong. His research interests include computer graphics and 3D computer visions.
\end{biography}
\vspace*{1.2em}

\begin{biography}[figures/authors/author5]{Ziyi Yang} received his bachelor’s degree from Shanghai Jiao Tong University in 2022. He is currently a research assistant in the CVMI lab at the University of Hong Kong, advised by Professor Xiaojuan Qi. His research interests include neural rendering and foundation models.
\end{biography}
\vspace*{1.2em}

\begin{biography}[figures/authors/author6]{Peng Dai} received the B.Eng. and M.Eng. degrees from the University of Electronic Science and Technology of China, and the Ph.D. degree from The University of Hong Kong. He is currently a Postdoctoral Fellow at the Vector Institute and the University of British Columbia. His research interests lie in neural rendering, generative artificial intelligence, world model, and immersive environment modeling and creation.
\end{biography}
\vspace*{1.2em}

\begin{biography}[figures/authors/author7]{Xiaojuan Qi} is an Associate Professor in the Department of Electrical and Computer Engineering at the University of Hong Kong. She received her Ph.D. from the Chinese University of Hong Kong and has conducted research and academic exchanges at the University of Toronto, the University of Oxford, and Intel’s Visual Computing Group. Her research focuses on empowering machines with the ability to perceive, understand, and reconstruct the visual world in open environments, with an emphasis on real-world deployment in embodied agents. She has authored over 100 papers in premier conferences across computer vision, graphics, and machine learning, including SIGGRAPH Asia, CVPR, ICCV, and NeurIPS, with many selected for oral presentations. Dr. Qi received a Best Paper Honorable Mention at SIGGRAPH Asia and was named one of IEEE AI’s 10 to Watch in 2024, MIT TR 35 China. She actively serves the academic community, frequently as an Area Chair for ICCV, CVPR, NeurIPS, ICML, and AAAI. 
\end{biography}

\vspace*{1.2em}
\subsection*{Graphical abstract}


\begin{figure*}[htbp]
    \centering
    \includegraphics[width=1.0\linewidth]{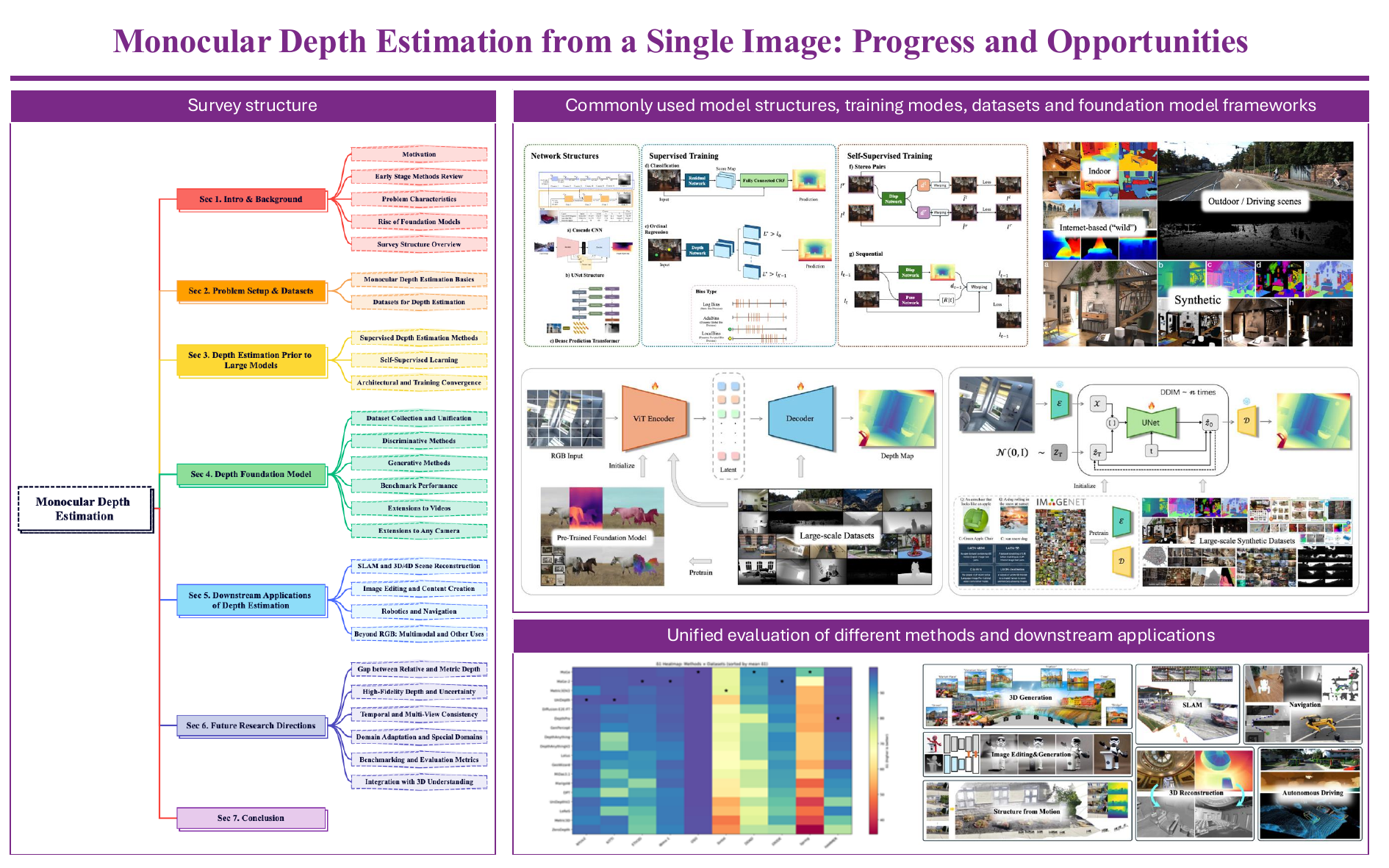}
    \caption{Graphical abstract}
\end{figure*}

\end{document}